\documentclass{fairmeta}

\usepackage[utf8]{inputenc} 
\usepackage{url}            
\usepackage{amsfonts}       
\usepackage{nicefrac}       
\usepackage{wrapfig}
\usepackage{makecell}
\usepackage{pifont}
\usepackage{amsmath}
\usepackage{xspace}

\usepackage{tikz}
\usetikzlibrary{arrows.meta, positioning, shapes.geometric, fit, backgrounds}
\usepackage{listings}
\usepackage{enumitem}

\definecolor{promptbg}{RGB}{248,248,248}
\definecolor{answerbg}{RGB}{244,248,255}
\definecolor{yamlbg}{RGB}{250,250,244}

\lstdefinestyle{promptstyle}{
  basicstyle=\ttfamily\scriptsize,
  frame=single,
  backgroundcolor=\color{promptbg},
  breaklines=true,
  columns=fullflexible,
  keepspaces=true,
  showstringspaces=false,
  tabsize=2
}

\lstdefinestyle{answerstyle}{
  basicstyle=\ttfamily\scriptsize,
  frame=single,
  backgroundcolor=\color{answerbg},
  breaklines=true,
  columns=fullflexible,
  keepspaces=true,
  showstringspaces=false,
  tabsize=2
}

\lstdefinestyle{yamlstyle}{
  basicstyle=\ttfamily\scriptsize,
  frame=single,
  backgroundcolor=\color{yamlbg},
  breaklines=true,
  columns=fullflexible,
  keepspaces=true,
  showstringspaces=false,
  tabsize=2
}

\newcommand{\aptbench}{\textsc{APT-Bench}}
\newcommand{\apt}{\textsc{APT}}

\AtBeginDocument{\hbadness=10000\hfuzz=200pt}

\title{{\setstretch{1.08}\fontsize{17}{20}\selectfont APT: Atomic Physical Transitions for Causal Video-Language Understanding\par}}

\author[1,*]{Shang Wu}
\author[1,*]{Haoran Lu}
\author[3,4]{Songling Liu}
\author[1]{Chenwei Xu}
\author[2]{Lie Lu}
\author[2]{Pranav Maneriker}
\author[2]{Fan Du}
\author[1]{Zhaoran Wang}
\author[1]{Han Liu}

\affiliation[1]{Northwestern~University}
\affiliation[2]{Dolby~Laboratories}
\affiliation[3]{Institute~of~Automation,~Chinese~Academy~of~Sciences,~Beijing~100190}
\affiliation[4]{University~of~Chinese~Academy~of~Sciences}

\contribution[*]{Equal contribution}

\abstract{
Physical events are not understood by their names alone, but by the causal state changes that compose them.
A clip-level label such as ``bounce'' can be correct while hiding the process that makes the event physically valid, from support loss and contact onset to rebound and settling.
To make this hidden process explicit, we introduce \textbf{Atomic Physical Transitions} (APTs): minimal, temporally localized state changes that bind a visible cue to an active physical mechanism and before/after dynamical regimes.
Our central contribution is APT as a \emph{structured transition interface} for mechanism-grounded physical video understanding: an APT chain represents a video as an ordered sequence of mechanism-typed, directed state changes rather than a single aggregate event label, so that
\textbf{event labels tell what happened; APT chains expose the causal structure of how it happened.}
To make APTs learnable by VLMs, we construct mixed-source APT data from human annotations and simulator ground truth, covering 14 transition types across contact, gravity, friction, and rotation/stability---the visually localizable macroscopic mechanics---with 27{,}303 timed instances over 1{,}246 trials.
Using this data, we find that current VLMs largely miss transition-level physics, with zero-shot recall at most 15\% and F1 at most 15\%, and errors dominated by missed transitions rather than timestamp jitter.
Direct fine-tuning on APT chains improves transition detection but causes event-level forgetting, indicating that the model learns a specialized answer format rather than a reusable physical representation.
We therefore propose \texttt{APT-Tune}, a parameter-efficient recipe that teaches VLMs to use causal transitions without forgetting how to answer ordinary video questions.
It combines image-pad-aware supervision, format-conditional co-training, and mechanism-conditioned domain-to-type decoding to make APT learning format-robust and physically grounded.
With only 11\,M LoRA parameters on Qwen3-VL-2B, \texttt{APT-Tune} substantially improves APT precision, recall, and F1 at a prediction budget matched to the ground truth, while also improving event-level video transfer within the macroscopic mechanics its taxonomy covers.
These results show that APTs are not a new answer format, but a human-aligned causal-structure supervision signal for physical video understanding. Project page: \url{https://polite-bombolone-d4250c.netlify.app/}
}

\begin{document}

\maketitle

\section{Introduction}
\label{sec:intro}

Recall a simple lesson from introductory mechanics: a ball bouncing on a table is not understood by the word ``bounce'' alone.
We understand the event by unpacking its causes: gravity accelerates the ball downward, contact with the table begins, collision forces reverse its motion.
This is how mechanics is typically organized in instruction: not as a list of event names, but as reusable principles such as force, contact, friction, collision, rotation, equilibrium, and gravitation~\citep{ling2016universityphysics,young2020university}.
Cognitive studies of physics expertise similarly show that experts represent problems by underlying physical principles rather than surface features~\citep{chi1981categorization,larkin1980expert}, while event cognition suggests that humans segment continuous activity at meaningful state boundaries~\citep{zacks2007event}.
These observations point to a causal view of physical video understanding: to understand an event, a model should recover when the active physical mechanism changes, why that transition occurs, and how it enables the next state.

Existing physics-oriented video benchmarks evaluate object interactions, prediction, counterfactual reasoning, and law-aware VQA or generation~\citep{yi2020clevrer,bear2021physion,tung2023physionpp,chow2025physbench,motamed2026physicsiq,zhang2025morpheus,wu2025mass,mak2026physicsmind}. However, most supervision remains tied to clip-level targets, such as answers, events, plausibility, or physical consistency. These targets reveal what happened but often leave the causal process underdetermined. We argue that physical video understanding should instead expose an ordered chain of temporally localized transitions, each linking a visible state change to its active mechanism. Such chains form a missing intermediate representation between event recognition and full physical simulation.

\begin{figure}[htbp]
    \centering
    \includegraphics[width=\linewidth]{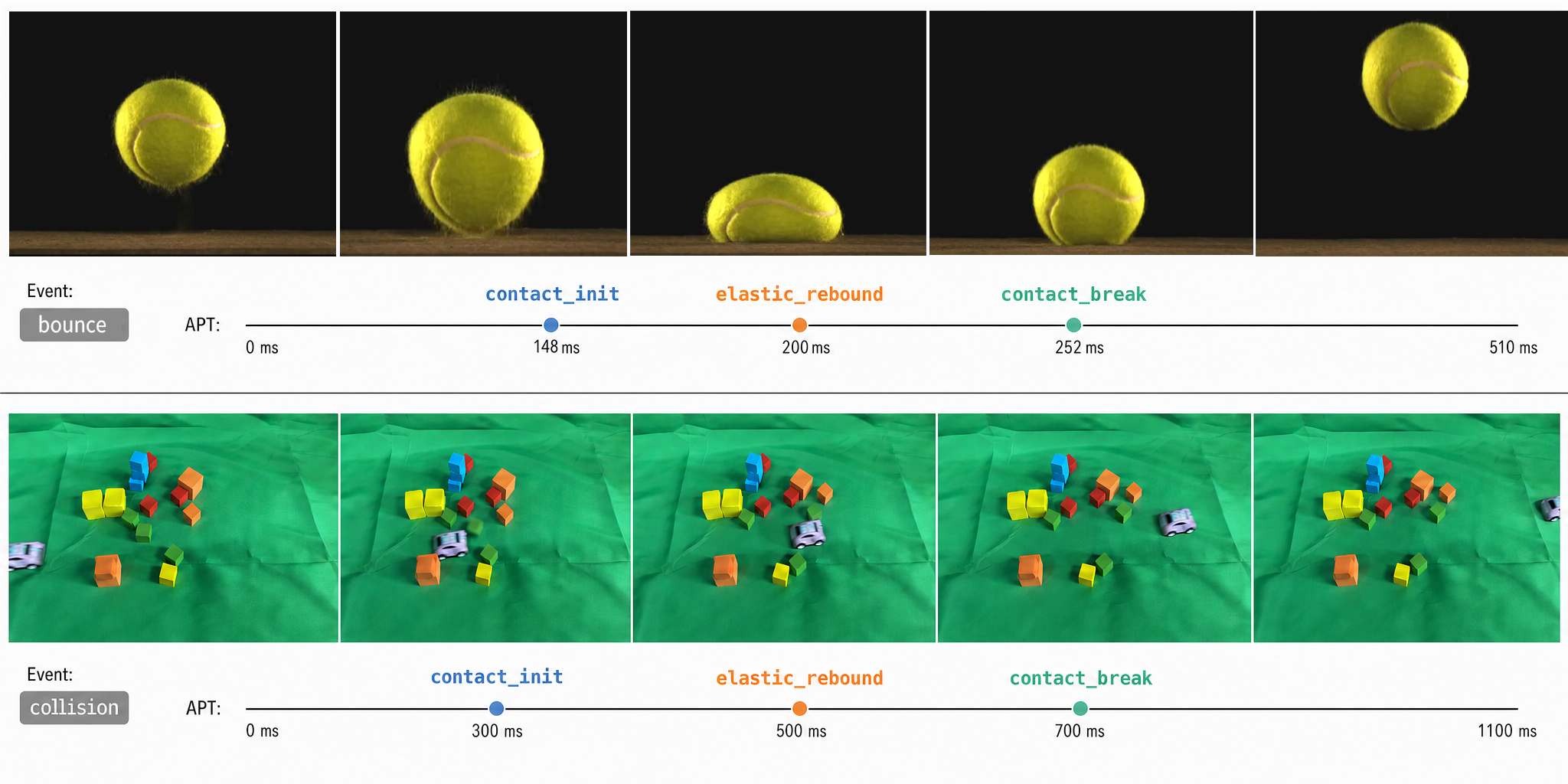}
    \caption{
    \textbf{Event labels say what happened; APT chains explain why.}
    A coarse event label such as \emph{bounce} or \emph{collision} summarizes a clip with a single outcome-level description, but hides the causal process that makes the event physically valid.
    Atomic Physical Transitions (APTs) decompose each clip into an ordered chain of timestamped, mechanism-typed state changes, such as \texttt{contact\_init}, \texttt{elastic\_rebound}, and \texttt{contact\_break}.
    Across visually different contact-driven scenes, APT chains expose the shared causal structure behind related event labels, providing a domain-agnostic supervision unit for transition-level physical understanding.
    }
    \label{fig:hero}
\end{figure}

To operationalize this view, we introduce \textbf{Atomic Physical Transitions} (APTs): minimal, visually grounded state changes in which a physical system moves between dynamical regimes.
APTs are not merely finer event labels, but causal reasoning units that bind a visible cue to an active mechanism and a before/after state.
A video is therefore represented as an ordered \emph{APT chain}, i.e., typed timestamps that recover the causal process hidden inside a clip-level event label.
Because APTs are defined by physical mechanism rather than scene appearance, the same schema can support annotation, evaluation, simulation specification, and training supervision.
We state the scope of this claim precisely: an APT type is a canonical name for a mechanism--state-change pair, so recovering an APT chain is \emph{causal-structure recovery}---identifying which mechanism becomes active, in which direction the state changes, and when---rather than intervention or counterfactual reasoning, which we do not evaluate.
Our present taxonomy is likewise scoped to visually localizable macroscopic mechanics and does not cover chemical, thermal, optical, or electromagnetic change.

APT-level supervision is hard to obtain by passive annotation alone, since clips may contain multiple brief transitions whose boundaries require physical judgment. We therefore build a mixed-source, human-calibrated pipeline. Human annotators label real-world recordings including CLEVRER~\citep{yi2020clevrer} to anchor the 14-label semantics, while simulator ground truth from Physion++ and Phys4D-style Isaac Sim scenes scales the data~\citep{tung2023physionpp,threedworld2021,phys4d2026,nvidia2025isaacsim, lu2024unigarment,lu2024garmentlab}. For simulated clips, poses, velocities, contacts, and support states provide structured evidence for candidate APTs; LLM/VLM prompting only maps this evidence into the shared schema and training formats, rather than inferring labels from pixels. Human validation checks this conversion, and source/render diversity reduces visual shortcuts.

Using this mixed-source APT data, we first test whether current VLMs can recover causal transition chains. They cannot: with the same 16-frame input and APT-schema prompt---each frame preceded by an explicit text marker giving its time in milliseconds---thirteen zero-shot VLMs, including GPT-5, Gemini 3.1, and Claude Sonnet 5, achieve only 10--15\% recall and at most 15\% F1 in the full 14-type setting, mainly due to missed transitions rather than timestamp jitter. This gap is not an artifact of sparse sampling: restricting predictions to the 16 provided frame times would already permit 61.8\% recall at the $\pm200$\,ms tolerance. Clip-level event competence therefore does not imply transition-level physical understanding. APT-only SFT improves recall but creates an APT-format specialist: because APT targets are structured chains while ordinary video tasks use QA, multiple choice, or short descriptions, the model overfits to the schema and loses prompt-conditioned event-level competence. The key challenge is therefore to make transition-level causality reusable across answer formats.

We address this with \texttt{APT-Tune}, a parameter-efficient SFT recipe that treats APT enumeration as an atomic causal-representation pretext task.
\texttt{APT-Tune} uses format-conditional co-training over APT, multiple-choice, and description formats, so the same video representation supports both local transition prediction and ordinary event-level answering.
Non-APT formats provide event-level rehearsal, while image-pad-aware masking focuses supervised loss on the structured answer rather than multimodal padding artifacts.
To encourage physical causality rather than label memorization, mechanism-conditioned APT prompts guide domain-to-type decoding, e.g., support loss under gravity to \texttt{free\_fall\_onset}, gap closure to \texttt{contact\_init}, post-contact velocity reversal to \texttt{elastic\_rebound}, and dissipated sliding to \texttt{sliding\_arrest}.
With only 11\,M LoRA parameters on Qwen3-VL-2B, \texttt{APT-Tune} raises APT recall from 10.0\% to 38.1\% (a 28.1\,pp gain) and F1 from 14.8\% to 38.8\%, while emitting 21.5 predictions per clip against a ground-truth mean of 22.3, so the gain reflects better prediction rather than overprediction.
It transfers to PhysBench without training on it and improves Something-Something\,v2 multiple-choice accuracy by 15\,pp over the base model.
These results show that APTs expose missing transition-level physics and provide format-robust causal-structure supervision for broader video understanding.

\noindent\textbf{Contributions.}
Our primary contribution is the APT representation itself; \aptbench{} is its diagnostic realization, the data pipeline makes the supervision scalable, and \texttt{APT-Tune} is its learning realization.
Concretely:
\begin{itemize}[leftmargin=*,topsep=1pt,itemsep=0pt,parsep=0pt,partopsep=0pt]
    \item \textbf{We propose APTs as a human-inspired causal transition representation.}
    Atomic Physical Transitions represent videos as ordered chains of visually grounded physical state changes, exposing both when a regime changes and why it changes.
    \item \textbf{We construct mixed-source APT data for diagnosis and supervision.}
    Our data combines human annotations, simulator ground truth, and diverse renderers, revealing that current VLMs miss most transition-level physics even when clip-level events are recognizable.
    \item \textbf{We introduce \texttt{APT-Tune} for format-robust causal fine-tuning.}
    \texttt{APT-Tune} learns APTs as causal representations rather than a JSON output style, addressing format mismatch, event-level forgetting, and label memorization.
\end{itemize}

\section{Related Work}
\label{sec:related_work}
\providecommand{\cmark}{\ding{51}}
\providecommand{\xmark}{\ding{55}}

Physics-oriented benchmarks evaluate causal video QA, physical prediction, VLM reasoning, and law- or motion-aware understanding~\citep{yi2020clevrer,bear2021physion,tung2023physionpp,chow2025physbench,motamed2026physicsiq,zhang2025morpheus}, but their targets are mostly clip-level answers, outcomes, plausibility judgments, QA labels, or trajectories.
APT is complementary: it makes localized, mechanism-typed state changes---support loss, contact initiation, rebound, sliding arrest, and settling---explicit targets for annotation, evaluation, and supervision.
This connects to event segmentation, physics expertise, and temporal video understanding: humans segment activity at meaningful boundaries, experts organize physics by underlying principles, and video methods model temporal structure through actions, interactions, or procedural steps~\citep{zacks2007event,chi1981categorization,larkin1980expert,goyal2017something,heilbron2015activitynet,damen2018epic,lea2017tcn,farha2019mstcn}.
APT instead treats boundaries as physical regime transitions, motivating \texttt{APT-Tune}: transition-level causal supervision with format-conditional co-training and parameter-efficient adaptation~\citep{liu2023visual,lin2024videollava,hu2022lora} to transfer across QA and description formats.

\paragraph{Concurrent work.}
Concurrently, \citet{groundingphysicalsignals2026} also argue that physical video understanding requires grounding rather than event naming alone, but at a different level of structure: they extend the what--when--where evaluation of V-STaR~\citep{cheng2025vstar} to 1{,}560 clips across six physics domains, scoring a \emph{single} event via an event description, a temporal interval, and a spatial box, and diagnose models through prompt families and input perturbations such as frame shuffling and masking.
APT instead requires recovering an unknown-length \emph{ordered chain} of mechanism-typed, directed state transitions, each localized by a canonical timestamp, and additionally uses that representation as training supervision in \texttt{APT-Tune} while measuring transfer to broader video understanding.
The two are complementary: single-event spatio-temporal grounding and chain-level mechanism recovery probe different aspects of the same gap, and their finding that spatial grounding is weakest aligns with our observation that transition \emph{recovery}, not localization, is the binding constraint.
A full discussion is given in Appendix~\ref{app:related_work}.

\begin{table*}[htbp]
\centering
\caption{\textbf{Comparison of evaluation units across physics and temporal video benchmarks.}
APT differs from prior benchmarks by making timestamped, mechanism-typed physical transitions the supervision and scoring unit, enabling ordered chain-level evaluation.}
\vspace{3mm}
\label{tab:apt_benchmark_comparison}
\small
\setlength{\tabcolsep}{3.8pt}
\renewcommand{\arraystretch}{1.45}
\resizebox{\textwidth}{!}{%
\begin{tabular}{@{}l l l l l l l l@{}}
\toprule
& \multicolumn{2}{c}{\textbf{Evaluation unit}}
& \multicolumn{3}{c}{\textbf{Physical / temporal structure}}
& \multicolumn{2}{c}{\textbf{Provenance}} \\
\cmidrule(lr){2-3}
\cmidrule(lr){4-6}
\cmidrule(lr){7-8}
\textbf{Data / Eval.} &
\textbf{Target} &
\textbf{Granularity} &
\makecell[c]{\textbf{Temporal}\\\textbf{Location}}&
\makecell[c]{\textbf{Physics}\\\textbf{Taxonomy}}&
\makecell[c]{\textbf{Ordered}\\\textbf{Chain}} &
\makecell[c]{\textbf{Label /}\\\textbf{Source}} &
\textbf{Scale (native)} \\
\midrule

CLEVRER~\cite{yi2020clevrer} &
Causal video QA &
Event QA &
\xmark &
Collision events &
Partial &
Sim GT &
20K vids / $>$300K QA \\

CLEVRER-Humans~\cite{Mao2023CLEVRERHumansDP} &
Causal event QA &
Event / edge QA &
\xmark &
Free-form events &
Graph &
Human on sim &
1,108 vids / 1,076 QA \\

Physion~\cite{bear2021physion} &
Physical prediction &
Future outcome &
Endpoint &
Object dynamics &
\xmark &
Sim GT + human &
1.2K examples / 8 scen. \\

Physion++~\cite{tung2023physionpp} &
Physical prediction &
Future outcome &
Endpoint &
Latent properties &
\xmark &
Sim GT + human &
8K train / 768 test$^\ddagger$ \\

PhysBench~\cite{chow2025physbench} &
Physical VLM QA &
Entry QA &
\xmark &
Broad domains &
\xmark &
Unspecified$^\dagger$ &
10,002 entries \\

Physics-IQ~\cite{motamed2026physicsiq} &
Video continuation &
Future continuation &
Partial &
Physical laws &
\xmark &
Real video GT &
396 vids / 66 scen. \\

MASS-Bench~\cite{wu2025mass} &
Physics video QA &
Entity motion &
Partial &
Physics categories &
Partial &
Human+AIGC &
4,350 vids / 8,361 QA \\

TemporalBench~\cite{cai2024temporalbench} &
Temporal QA / caption &
Sub-event QA &
Partial &
Temporal only &
Partial &
Human + LLM &
10K QA / 2K ann. \\

\midrule
\textbf{APT (ours)} &
\textbf{APT chain recovery} &
\textbf{Transition chain} &
\textbf{Timestamp} &
\textbf{14 APTs / 4 domains} &
\textbf{\cmark} &
\textbf{Human + Sim GT} &
\textbf{27.3K APTs / 1,246 trials} \\
\bottomrule
\end{tabular}%
}
\end{table*}

\section{Atomic Physical Transitions: From Event Labels to Causal Chains}
\label{sec:apt}
\subsection{Events as Outcomes, APTs as Causes}
\label{sec:apt_intuition}
A clip-level event label names what happened, but collapses the causal process that makes it happen: a ``bounce'' may include contact initiation, collision response, contact break, and settling. This matches how mechanics is learned: reusable physical processes such as gravity, contact, collision, friction, rotation, and stability are learned separately, while real events compose them over time. We therefore define Atomic Physical Transitions (APTs) by \emph{physical process}, not by object or material category; rigid objects, soft bodies, granular materials, and fluid-like regions can instantiate the same APT when the active mechanism and visible before/after state change are the same. Event labels tell what happened; APT chains expose the causal structure of how it happened. Detailed motivation is given in Appendix~\ref{app:apt_motivation}.

\subsection{Definition}
\label{sec:apt_definition}

\noindent\textbf{Definition.}
Given a video $V$ of a physical scene, an \textbf{Atomic Physical Transition} is a visually localizable moment at which a physical entity, material region, or interacting pair changes between two coarse physical states under a specific active mechanism.
We represent an APT as
\begin{equation}
    a_i = (\tau_i, d_i, \ell_i, s_i^- \rightarrow s_i^+, e_i),
    \label{eq:apt_def}
\end{equation}
where $\tau_i$ is the canonical transition time, $d_i$ is the physical domain, $\ell_i \in \mathcal{L}_{d_i}$ is the transition type, $s_i^- \rightarrow s_i^+$ is the before/after physical state change, and $e_i$ is the local visual evidence supporting the transition. The domain $d_i$ groups transitions by active physical process rather than by object class or material type; the type $\ell_i$ names the specific state change within that process family; and the evidence $e_i$ anchors the label to local visual cues. Thus material affects how an APT appears, but not what the APT means: an APT records not only \emph{when} a state changes, but also \emph{why} it changes.

Atomicity is defined relative to our vocabulary and video timescale. An APT is atomic if, for the same entity or interacting pair, the local neighborhood of $\tau_i$ contains no other named APT that further explains the same transition $s_i^- \rightarrow s_i^+$. In this sense, APTs are not physically indivisible events, but the smallest causal units we ask models to localize, classify, and reuse across answer formats. Granularity is therefore set by physical meaning and visual observability rather than by subdividing the vocabulary arbitrarily: a transition is admitted only if it is physically meaningful, supported by a local neighborhood of at least three frames, and atomic under the current vocabulary. Thus \texttt{sliding\_arrest} marks the visible end of translational sliding, while residual wobble that later disappears is labeled \texttt{settling} instead of being decomposed into individual oscillations; sub-frame deformation and force oscillations need a different capture timescale and lie outside the benchmark.

\paragraph{What scoring $(\tau_i,\ell_i)$ measures.}
Although Eq.~\eqref{eq:apt_def} names a mechanism and a directed state change, evaluation scores only type and timestamp. This target is stronger than it appears, because each label $\ell_i$ is a \emph{canonical name} for a mechanism--state-change pair: \texttt{free\_fall\_onset} denotes support loss followed by supported$\rightarrow$falling dynamics, and \texttt{elastic\_rebound} denotes collision response followed by approach$\rightarrow$retreat dynamics. Scoring $(\tau_i,\ell_i)$ therefore measures recovery of this canonical causal structure rather than free-form mechanism explanation. We accordingly claim causal-\emph{structure} recovery, not general causal reasoning: \aptbench{} does not test intervention or counterfactual queries. Chains are further constrained by physically admissible composition---support loss enables falling, falling enables contact, contact activates collision response---so predictions are scored against a causally coherent reference rather than an arbitrary label set.

\begin{figure*}[htbp]
    \centering
    \includegraphics[width=\linewidth]{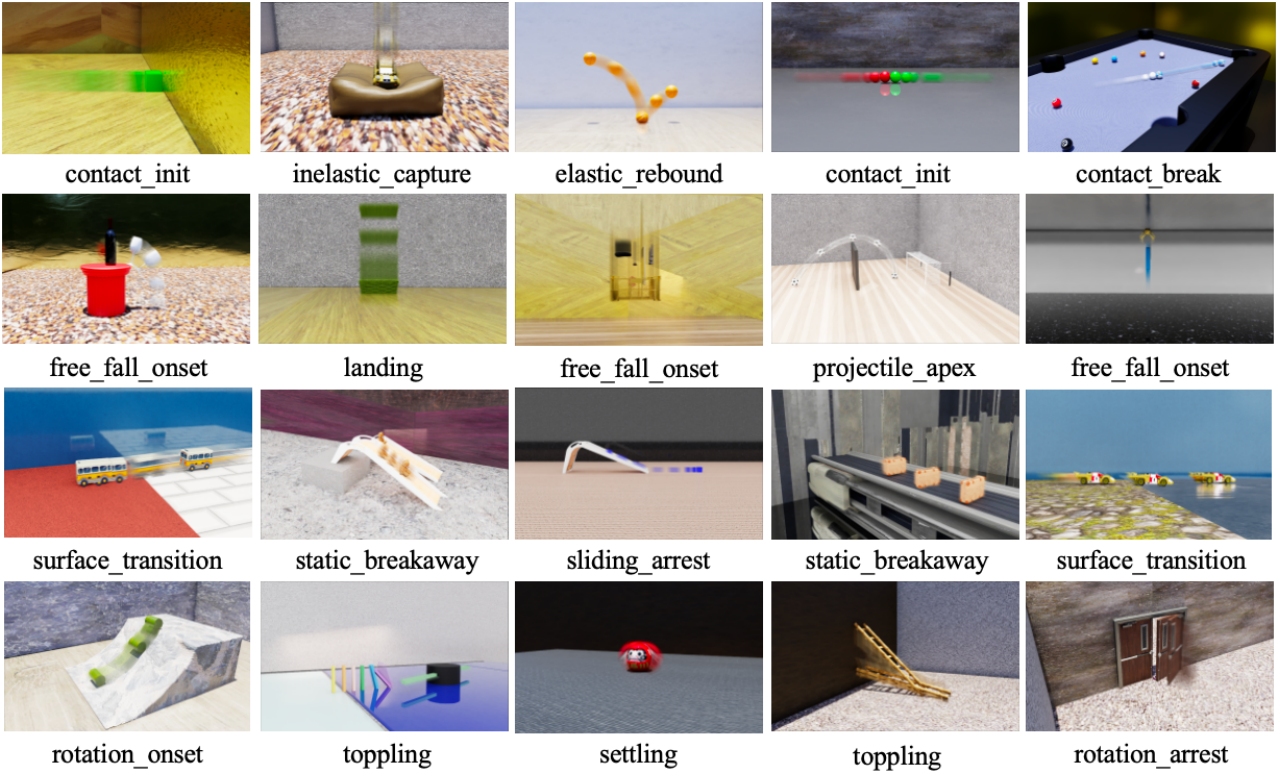}
    \caption{Event-level comparison between standard event-level evaluation and APT.}
    \label{fig:apts}
\end{figure*}

\subsection{Problem Formulation: APT Chain Enumeration}
\label{sec:apt_formulation}
Let $V=(I_1,\ldots,I_T)$ be a video with duration $T$. Transition-level physical understanding requires recovering an ordered chain of APTs:
\begin{equation}
    \mathbf{A}(V) = (a_1,\ldots,a_K), \quad
    0 \leq \tau_1 \leq \cdots \leq \tau_K \leq T,
    \qquad
    a_i^{\mathrm{eval}} = (\tau_i, \ell_i),
    \label{eq:apt_chain}
\end{equation}
where $K$ is unknown and varies by video, and each evaluation item is serialized as a typed timestamp. The task is therefore to recover how many transitions occur, when they occur, and which physical type each transition has.
\emph{Why timestamps rather than frame indices or order alone.} Sequence order records only which transition precedes which, whereas a timestamp additionally binds each transition to its local visual evidence, separates repeated instances of the same type, and gives a common coordinate across the differing frame rates and durations of our sources (Table~\ref{tab:source_stats}); a frame index is equivalent only under a single fixed frame rate and sampling scheme. This is not a reduction to frame classification: the model must still predict the unknown number, types, instances, and order of transitions, and the contribution is the ordered mechanism-grounded chain rather than any particular temporal coordinate. An APT chain is a causal representation, not a fixed output style: the same chain can be serialized as JSON, verbalized as a description, or queried through QA prompts. This distinction matters because treating APT supervision as a JSON template can produce an APT-format specialist, whereas learning it as a shared causal representation supports both transition enumeration and ordinary event understanding. We therefore formulate APT prediction as mechanism-conditioned decoding: first identify the active physical domain, then decode the transition type.

\subsection{APT Taxonomy}
\label{sec:apt_taxonomy}

Our scope is macroscopic, visually grounded physical dynamics across rigid, deformable, granular, and fluid scenes.
APTs are process-centered: labels are determined by the active mechanism and visible before/after state change, not by object or material category.
They cover localizable transitions such as contact onset/loss, support loss, collision response, friction- or medium-mediated motion change, rotation onset/arrest, stability loss, and settling, while excluding invisible changes, continuous parameter estimation, or full physical-state reconstruction.
Labels are assigned to the entity, region, or interacting pair undergoing the transition.
The 14 labels are \emph{mutually exclusive per transition instance}: each annotated instance receives exactly one most-specific label, resolved by the active mechanism, the before/after state, and a specific-over-generic rule.
Distinct transitions may nevertheless share a timestamp when they describe different state variables---\texttt{contact\_init} (separated$\rightarrow$interacting), \texttt{landing} (falling$\rightarrow$supported), and \texttt{inelastic\_capture} (moving$\rightarrow$captured/resting) can co-occur at one impact---so apparent overlap reflects concurrent independent changes, not ambiguous labels.
The taxonomy is exhaustive only within the stated scope of visually grounded macroscopic mechanics, and is not claimed to be exhaustive over physics as a whole.
Because an APT is defined locally, a scene-level chain naturally interleaves transitions from several entities and relations, and the present data already contain multi-object processes such as concurrent falling, landing, contact, and settling; that said, most clips contain 2--3 APT-relevant interacting objects (at most 5), so dense multi-object dynamics with simultaneous contacts and severe occlusion remain under-represented (Appendix~\ref{app:limitations}).
Ambiguous cases use the label that best explains the active mechanism, preferring specific transitions over generic ones; detailed boundary rules are in Appendix~\ref{app:apt_boundary_rules}.
\begin{table*}[htbp]
\centering
\caption{\textbf{APT taxonomy.}
APTs are organized by physical process domain and defined by coarse before/after state changes.}
\label{tab:taxonomy}
\footnotesize
\setlength{\tabcolsep}{4pt}
\begin{tabular}{@{}p{2.7cm}p{1.7cm}p{1.7cm}@{\hspace{1.2em}}p{2.7cm}p{1.7cm}p{1.7cm}@{}}
\toprule
\multicolumn{3}{c}{\textbf{Domain I: Contact \& Collision}} &
\multicolumn{3}{c}{\textbf{Domain IV: Rotation \& Stability}} \\
\cmidrule(r){1-3}\cmidrule(l){4-6}
\textbf{Label} & \textbf{Before} & \textbf{After} &
\textbf{Label} & \textbf{Before} & \textbf{After} \\
\midrule
\texttt{contact\_init} & separated & interacting &
\texttt{rotation\_onset} & non-rotating & rotating \\
\texttt{contact\_break} & interacting & separated &
\texttt{toppling} & stable & unstable \\
\texttt{elastic\_rebound} & approach & retreat &
\texttt{rotation\_arrest} & rotating & stationary \\
\texttt{inelastic\_capture} & moving & captured/rest &
\texttt{settling} & disturbed & equilibrium \\
\midrule
\multicolumn{3}{c}{\textbf{Domain II: Gravity \& Projectile}} &
\multicolumn{3}{c}{\textbf{Domain III: Friction, Surface \& Medium}} \\
\cmidrule(r){1-3}\cmidrule(l){4-6}
\textbf{Label} & \textbf{Before} & \textbf{After} &
\textbf{Label} & \textbf{Before} & \textbf{After} \\
\midrule
\texttt{free\_fall\_onset} & supported & falling &
\texttt{static\_breakaway} & stationary & moving \\
\texttt{projectile\_apex} & ascending & descending &
\texttt{sliding\_arrest} & moving & stationary \\
\texttt{landing} & falling & supported &
\texttt{surface\_transition} & surface A & surface B \\
\bottomrule
\end{tabular}
\end{table*}

Our taxonomy contains 14 recurring transitions grouped by active physical process.
The same APT can be instantiated by rigid objects, soft bodies, granular materials, or fluid-like regions when the causal mechanism and before/after state change match.
\textbf{Contact \& collision} captures interaction-boundary changes and outcomes;
\textbf{gravity \& projectile} captures support loss, apex reversal, and landing;
\textbf{friction, surface \& medium} captures rest-to-motion, motion-to-rest, and surface/medium changes;
and \textbf{rotation \& stability} captures angular-motion changes, toppling, and final equilibrium.

\section{Rule-Guided Compositional APT Data Construction}
\label{sec:benchmark}

APT supervision is hard to collect because a dynamic chain must localize fine-grained causal transitions within a physically plausible composition; we therefore use a rule-guided coarse-to-fine pipeline.

\begin{figure}[htbp]
    \centering
    \includegraphics[width=\linewidth]{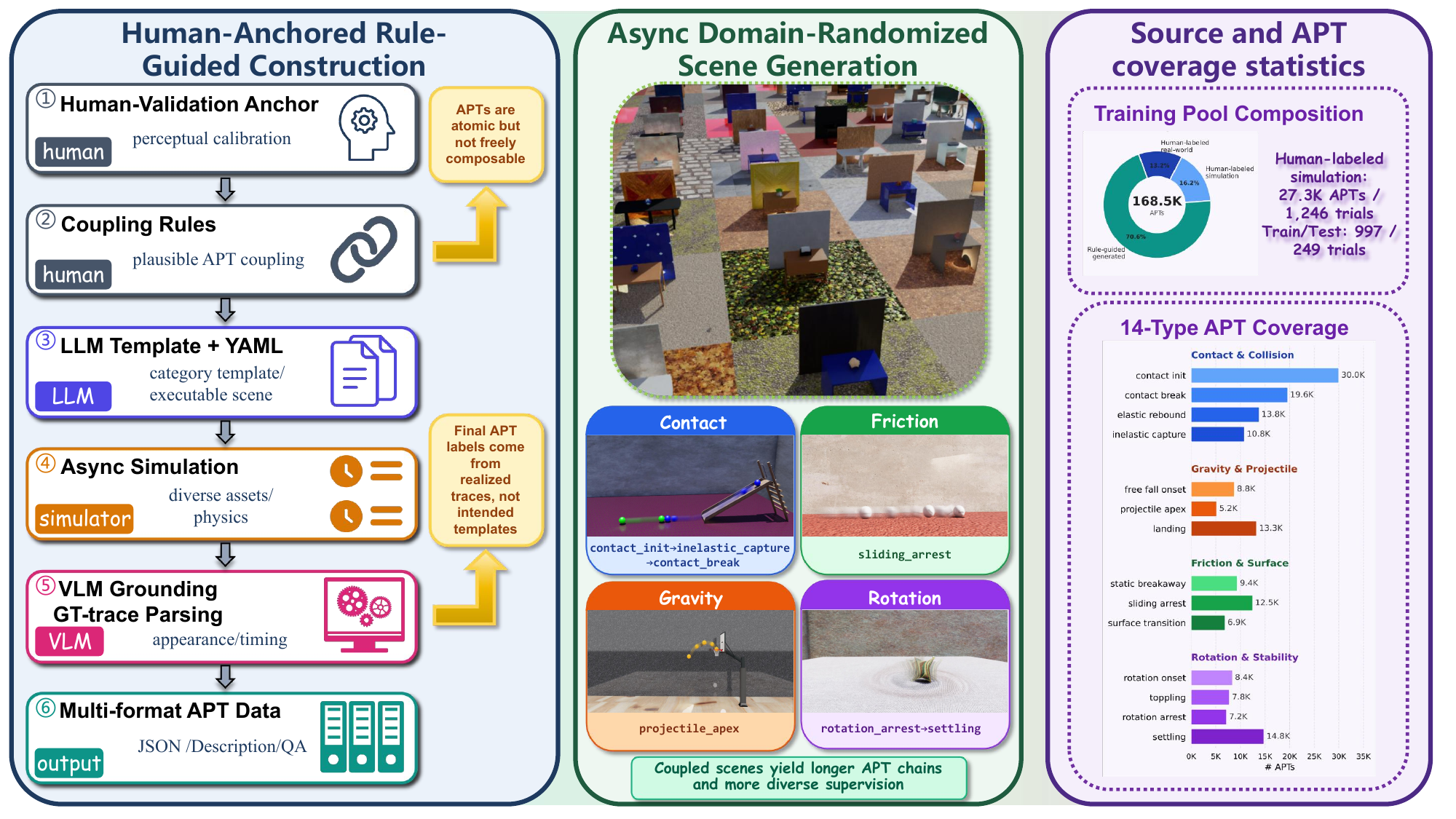}
    \caption{
\textbf{APT data construction pipeline.}
Human validation anchors APT semantics and coupling rules; LLM/YAML templates drive asynchronous domain-randomized simulation; VLM grounding identifies rendered objects; and simulator GT traces provide realized APT labels and timestamps.
The accepted clips are exported as multi-format supervision with full 14-type coverage.
}
    \label{fig:pipeline}
\end{figure}

\subsection{Human-Labeled Validation Anchor}
\label{sec:human_validation}

We use human annotation as the perceptual anchor for APT construction, since APT boundaries must align with visible physical transitions rather than simulator states alone.
Five trained annotators, all following the same 14-type guidelines, label controlled videos from CLEVRER~\citep{yi2020clevrer} and Physion++~\citep{tung2023physionpp} by marking visible state changes, assigning one of 14 APT types, and recording timestamps.
Every video is annotated independently by two annotators for transition segmentation, type, and timestamp, and any disagreement is adjudicated by a third annotator before the trial is admitted.
This yields 1{,}246 human-labeled trials with 27{,}303 APTs: 500 CLEVRER trials with 6{,}316 APTs covering 7 types, and 746 Physion++ trials with 20{,}987 APTs covering the full taxonomy.
These labels serve as calibration anchors, not the scaling mechanism, aligning templates, prompts, and GT-trace parsing with human-visible transitions.
Two reliability caveats follow.
Because adjudication preceded record commitment, we did not retain the pre-adjudication labels needed for an inter-annotator agreement coefficient, so we report the procedure rather than a $\kappa$ statistic; and the GT-trace parser, used only for simulator-derived \emph{training} data and never for the human-labeled test ground truth, was calibrated against the human anchor rather than evaluated held-out, so no human-versus-parser agreement number is reported.

\subsection{Rule-Guided Coarse-to-Fine Simulation}
\label{sec:rule_guided_sim}

We scale APT data with a rule-guided coarse-to-fine simulation pipeline~\citep{phys4d2026} that assigns each component its natural role: humans provide physical commonsense, LLMs expand structured templates, coding agents make them executable, simulators provide GT traces, and VLMs ground rendered appearance.
The pipeline starts from \emph{human-defined coupling rules}, since APTs are atomic but not freely composable: gravity naturally couples with falling, landing, contact, rebound, and settling; friction with sliding and arrest; and stability loss with rotation, toppling, and settling.
Given a coupling rule, an LLM proposes category-level scene templates under simulator asset constraints; a coding agent converts each template into executable YAML with object slots, assets, initial states, physical parameters, cameras, and success checks.
We then run asynchronous domain-randomized simulation, instantiating diverse objects, colors, materials, dynamics, and initial conditions while preserving the intended causal coupling.

Rendered videos and simulator traces are refined into instance-specific APT data.
Because concrete assets are sampled only at simulation time, a VLM grounds the abstract template into visible objects, colors, materials, and relations.
Precise APT stages are not inferred from pixels: an LLM-assisted parser reads GT traces---pose, velocity, contact pairs, support relations, and object identity---and converts state changes into realized APT instances.
The intended template is only a prior; final APT chains are determined by realized traces, enabling filtering, relabeling, and retention of additional transitions caused by randomized dynamics.
We validate these automatic labels against the human anchor, using disagreements to refine coupling rules, YAML checks, grounding prompts, parsing rules, and motion/contact thresholds.

Each accepted clip is exported as aligned APT JSON, grounded description, causal transition description, and event-level QA/MCQ, preventing JSON-only collapse in mixed-format \texttt{APT-Tune}.

\subsection{Evaluation Protocol and Benchmark Statistics}
\label{sec:eval_protocol}

Evaluation uses a stratified split of 997 train trials ($\sim$21.8K APTs) and 249 test trials ($\sim$5.5K APTs), with 16 uniformly sampled frames, one shared APT-schema prompt, tolerant JSON parsing, and one-to-one type matching within $\pm 200$\,ms.
Because the two sources differ in duration and frame rate (Table~\ref{tab:source_stats}), we score timestamps in milliseconds rather than frame indices.

\paragraph{Frame times are given, not inferred.}
Every sampled frame is preceded by an explicit text marker, \texttt{[Frame i/16, t=...ms] <image>}, whose value is computed from the frame's native index and the clip's native frame rate; the prompt header additionally states the frame count, duration, native frame rate, and valid timestamp range.
This format is byte-identical across training, inference, and all backbones, and neither Qwen3-VL nor InternVL3.5 uses a family-specific video-timestamp interface, so every model receives equivalent temporal information and only the weights differ across rows of our result tables.
The model is thus asked to predict \emph{transition} times from visual evidence conditioned on known frame times, not to infer frame times from pixels.

\paragraph{Sparse sampling does not explain the low recall.}
Since 16 frames are spaced 320--806\,ms apart while the tolerance is $\pm 200$\,ms, we measure the \emph{frame-grid ceiling}: the recall attainable if predictions were restricted to the sampled-frame timestamps.
At 16 frames this ceiling is 61.8\%, far above the 10--15\% zero-shot results, so timestamp quantization is not the binding constraint.
Under an 8-frame budget tuned models reach 44.6--48.5\%, \emph{above} the corresponding 33.8\% ceiling, so they interpolate between provided frame times rather than copying them, and a tolerance sweep from $\pm 100$ to $\pm 800$\,ms preserves the model ordering throughout (Appendix~\ref{supp:frame_grid_ceiling}).
Sparse sampling can still hide transitions briefer than the frame spacing, a limitation of the current protocol.

\begin{table}[t]
\centering
\small
\caption{\textbf{\aptbench{} test-split statistics by source.} Frame spacing is the interval between uniformly sampled frames; the last column is the mean number of ground-truth APTs per clip. The two sources differ substantially in duration and frame rate, motivating millisecond-valued timestamps.}
\label{tab:source_stats}
\setlength{\tabcolsep}{5pt}
\begin{tabular}{@{}lcccccc@{}}
\toprule
\textbf{Source} & \textbf{\#Test} & \textbf{Duration (s)} & \textbf{FPS} & \textbf{Resolution} & \textbf{16-frame spacing} & \textbf{APTs/clip} \\
\midrule
CLEVRER   & 100 & 5.12                  & 25 & $480{\times}320$ & 320\,ms & 12.0 \\
Physion++ & 149 & 12.89 (7.00--21.87)   & 30 & $256{\times}256$ & 806\,ms & 29.2 \\
\midrule
\textbf{All} & \textbf{249} & \textbf{9.77} & --- & --- & --- & \textbf{22.3} \\
\bottomrule
\end{tabular}
\end{table}

\section{\texttt{APT-Tune}: Learning from APT Chains}

\begin{figure}[t]
    \centering
    \includegraphics[width=\linewidth]{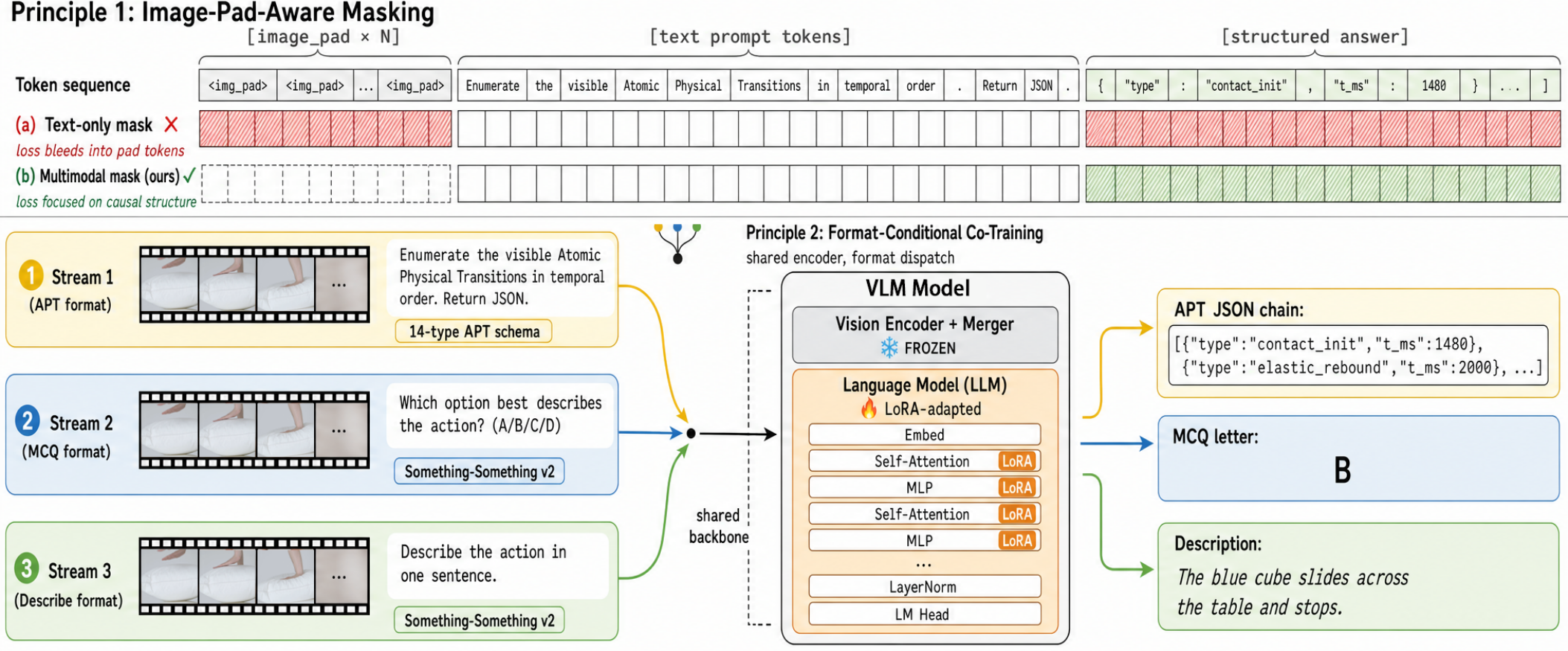}
    \caption{\textbf{\texttt{APT-Tune}: two design principles.} \emph{(1) Image-pad-aware masking.} The processor expands the image placeholder into hundreds of \texttt{<image\_pad>} tokens. A text-only mask (a) leaves them in the objective, while our multimodal-aware mask (b) excludes them and focuses loss on the structured answer. \emph{(2) Format-conditional co-training.} APT JSON, MCQ letter, and one-sentence description share a single Qwen3-VL backbone with a frozen vision encoder and merger and LoRA on the LLM, dispatched per record to prevent APT-JSON specialist collapse.}
    \label{fig:method}
\end{figure}
\label{sec:method}

\texttt{APT-Tune} is a parameter-efficient SFT recipe that treats APT enumeration as a pretext task and rests on two principles.

\paragraph{Principle 1: Image-pad-aware masking.}
The APT chain is a structured target where every token has a clear role (transition type, timestamp, or scaffold), so loss should flow only over the chain.
In multimodal SFT, the processor expands each \texttt{<image>} placeholder into a grid of \texttt{<|image\_pad|>} tokens, and a text-only prefix length leaves around 500 pad tokens unmasked, turning part of the objective into pad-token reconstruction.
We replace $L_\text{text}{=}\mathrm{len}(\mathrm{tokenizer}(\text{prompt}))$ with $L_\text{mm}{=}\mathrm{len}(\mathrm{processor}(\text{prompt}, \text{images}).\mathrm{input\_ids})$, so the loss is computed only over the answer.
We stress that this is a statement about the \emph{loss region}, not the input: the interleaved frame-time markers of Section~\ref{sec:eval_protocol} are ordinary text tokens that remain fully visible to the model in both training and inference, and masking never removes them from the context.
The two mechanisms therefore act at different stages---timestamp tokens condition the prediction, whereas image-pad-aware masking decides which \emph{target} tokens receive gradient---and Table~\ref{tab:ablation} shows the cost of conflating them: with the text-only prefix length, cross-entropy plateaus at 1.92 because the model is partly trained to reconstruct \texttt{<|image\_pad|>} tokens.

\paragraph{Principle 2: Format-conditional co-training.}
A specialist trained only on APT-JSON collapses $p(\text{format}\mid\text{prompt})$ onto APT-JSON and loses competence on other tasks (\emph{specialist collapse}).
\texttt{APT-Tune} therefore co-trains three formats on a single encoder under a shared chat template, dispatched by a record-level \texttt{format\_type}.
The mix is small and balanced: 997 APT records (\aptbench{} train + Isaac Lab generated) plus 1{,}000 MCQ and 1{,}000 one-sentence-description records drawn from SSv2~\citep{goyal2017something}, an action corpus with no overlap with our zero-shot splits.
The non-APT records are not label augmentation, but maintain the format-conditional posterior while the APT objective trains atomic video structure.

\section{Experiments}
\label{sec:experiments}

We answer four questions.
\textbf{(Q1)} Do current VLMs recover transition-level physics zero-shot?
\textbf{(Q2)} Does \texttt{APT-Tune} improve APT detection across families and scales?
\textbf{(Q3)} Does APT supervision transfer to event-level video (MVBench) and OOD physics (PhysBench)?
\textbf{(Q4)} Are both design principles necessary?
All \aptbench{} numbers use a frozen protocol with 16 frames, the 14-type APT-schema prompt, tolerant JSON parsing, and one-to-one matching by type within $\pm 200$\,ms on the 249-trial test split ($\sim$5{,}500 APTs).

\subsection{Zero-Shot Diagnostic}
\label{sec:zeroshot}

\paragraph{Metrics.}
We report one metric per sub-problem of APT chain enumeration.
\textbf{Recall} (primary) is the fraction of ground-truth APTs detected and typed within $\pm 200$\,ms.
\textbf{Precision} and \textbf{F1} are computed under the same one-to-one type matching, and we report the mean \textbf{number of predictions per clip} alongside them, so that coverage and correctness can be read separately.
\textbf{Type accuracy} on matched events isolates type confusion from missed detection.
\textbf{Median timing error} $\Delta T$ over matched events measures localization precision (median for robustness to JSON-parse outliers).
Earlier drafts of this work reported recall alone on the grounds that precision is sensitive to JSON formatting variance across closed and open APIs; we now report the full set, and note that the tolerant parser of Appendix~\ref{supp:evaluation_protocol} affects fewer than 3\% of predictions, so formatting variance does not drive the precision numbers.

\begin{table}[t]
\centering
\small
\caption{\textbf{Zero-shot \aptbench{} (\%).} Recall and type accuracy at $\pm 200$\,ms; $\Delta T$ in ms. No model exceeds 15.2\% recall, and scale within a family does not help.}
\label{tab:zeroshot}
\setlength{\tabcolsep}{5pt}
\begin{tabular}{@{}lccc@{\hskip 2.2em}lccc@{}}
\toprule
\textbf{Model} & \textbf{Rec.} & \textbf{Type} & \textbf{$\Delta T$} &
\textbf{Model} & \textbf{Rec.} & \textbf{Type} & \textbf{$\Delta T$} \\
\midrule
GPT-4.1            & 13.0 & 15.0 & 95  & Qwen3-VL-2B    & 10.0 & 12.0 & 100 \\
Gemini 2.5 Flash   & 12.0 & 14.0 & 110 & Qwen3-VL-4B    & 11.0 & 13.0 & 95  \\
GPT-5              & 13.6 & 22.0 & 67  & Qwen3-VL-8B    & 12.0 & 14.0 & 90  \\
Gemini 3.1         & 12.8 & 23.1 & 67  & InternVL3.5-2B & 14.0 & 16.0 & 67 \\
Claude Sonnet 5    & \textbf{15.2} & \textbf{25.6} & 67 & InternVL3.5-4B & 13.0 & 15.0 & 75  \\
VideoLLaMA-3-2B    & 12.4 & 10.5 & 95  & InternVL3.5-8B & 13.0 & 15.0 & 80  \\
VideoLLaMA-3-7B    & 13.9 & 13.4 & 80  &                &      &      &     \\
\bottomrule
\end{tabular}
\end{table}

\paragraph{Findings.}
We evaluate five commercial frontier models (GPT-4.1~\citep{gpt41}, GPT-5~\citep{gpt5}, Gemini 2.5 Flash~\citep{gemini2}, Gemini 3.1~\citep{gemini31}, Claude Sonnet 5~\citep{claudesonnet5}) and three open-source families (Qwen3-VL-\{2,4,8\}\,B~\citep{qwen3vl2025}, InternVL3.5-\{2,4,8\}\,B~\citep{internvl352025}, VideoLLaMA-3-\{2,7\}\,B~\citep{videollama3}) under the identical frozen protocol.
Recall stays between 10.0 and 15.2\% across all thirteen systems (Table~\ref{tab:zeroshot}), with the strongest, Claude Sonnet 5, reaching only 15.2\%.
Among matched events, $\Delta T$ falls between 67 and 110\,ms, well within the $\pm 200$\,ms tolerance, so the gap reflects missed transitions rather than timing jitter.
Type accuracy stays between 10.5 and 25.6\%; the newest frontier models improve type accuracy on the events they do detect (up to 25.6\%) without improving recall, so their advantage lies in naming transitions rather than finding them.
Scaling open-source backbones from 2\,B to 8\,B yields no improvement, so the bottleneck is representation rather than capacity.
We scope this conclusion to the thirteen evaluated models rather than to VLMs in general.

\subsection{\texttt{APT-Tune} Across Six Open-Source Backbones}
\label{sec:main_results}

\paragraph{Recipe.}
We instantiate \texttt{APT-Tune} on all six open-source models from Section~\ref{sec:zeroshot}.
For each backbone we freeze the vision encoder and merger and apply LoRA~\citep{hu2022lora} ($r{=}16,\,\alpha{=}32$) to the LLM attention and MLP projections, training $\sim$11\,M parameters (0.5\% of the model) for under a day on a single A100 with bf16 AdamW at lr $1{\times}10^{-4}$, cosine schedule, 30 warm-up steps.
APT-format records come from the \aptbench{} train split and the Isaac Lab~\citep{mittal2025isaaclab} APT-style pool. MCQ and describe records come from SSv2~\citep{goyal2017something}.
The same hyperparameters and data mix are used on every backbone with no per-model tuning.

\begin{table}[t]
\centering
\small
\caption{\textbf{\texttt{APT-Tune} on six open-source backbones, evaluated on \aptbench{}.} Recall and type accuracy (\%) are matched at $\pm 200$\,ms; $\Delta T$ in ms; $\Delta$Rec is the recall gain over zero-shot. Best per column in bold.}
\label{tab:apt_tune}
\setlength{\tabcolsep}{6pt}
\begin{tabular}{@{}lccc@{\hskip 1em}ccc@{\hskip 1em}c@{}}
\toprule
& \multicolumn{3}{c}{\textbf{Zero-shot}} & \multicolumn{3}{c}{\textbf{+\texttt{APT-Tune}}} & \\
\cmidrule(lr){2-4} \cmidrule(lr){5-7}
\textbf{Backbone} & \textbf{Rec.} & \textbf{Type} & \textbf{$\Delta T$} & \textbf{Rec.} & \textbf{Type} & \textbf{$\Delta T$} & \textbf{$\Delta$Rec.} \\
\midrule
Qwen3-VL-2B    & 10.0          & 12.0          & 100          & 38.1          & 52.1          & 38          & 28.1 \\
Qwen3-VL-4B    & 11.0          & 13.0          & 95           & 42.5          & 57.8          & 33          & 31.5 \\
Qwen3-VL-8B    & 12.0          & 14.0          & 90           & 48.6          & 66.2          & 27          & 36.6 \\
\midrule
InternVL3.5-2B & \textbf{14.0} & \textbf{16.0} & \textbf{67}  & 46.8          & 61.5          & 25          & 32.8 \\
InternVL3.5-4B & 13.0          & 15.0          & 75           & 48.0          & 63.5          & 23          & 35.0 \\
InternVL3.5-8B & 13.0          & 15.0          & 80           & \textbf{53.4} & \textbf{71.2} & \textbf{19} & \textbf{40.4} \\
\bottomrule
\end{tabular}
\end{table}

\paragraph{Results.}
\texttt{APT-Tune} closes most of the zero-shot gap on every backbone with the same recipe (Table~\ref{tab:apt_tune}).
Qwen3-VL gains 28.1, 31.5, and 36.6\,pp recall at 2, 4, and 8\,B, and InternVL3.5 gains 32.8, 35.0, and 40.4\,pp at the same scales.
The best post-tune model (InternVL3.5-8B) reaches \textbf{53.4\%} recall, roughly $4\times$ its 13.0\% zero-shot baseline, and even the smallest fine-tuned model (Qwen3-VL-2B at 38.1\%) exceeds every zero-shot system by more than 24\,pp.
Type accuracy rises four- to five-fold, going from between 12 and 16\% pre-tune to between 52 and 71\% post-tune, and $\Delta T$ roughly halves, going from between 67 and 100\,ms pre-tune to between 19 and 38\,ms post-tune.

\paragraph{The recall gain is not overprediction.}
Because the task does not cap the number of predicted transitions, a recall gain could in principle come from emitting more events rather than better ones; Table~\ref{tab:prf} rules this out on Qwen3-VL-2B.
One-to-one matching already stops duplicate predictions from repeatedly matching the same ground-truth transition, but a larger output set can still raise recall, so we check the budget directly.
\texttt{APT-Tune} emits 21.5 predictions per clip against a ground-truth mean of 22.3, versus 7.7 zero-shot---moving \emph{toward} the correct output cardinality, not past it---while precision rises from 28.8 to 39.6\% and F1 from 14.8 to 38.8\%.
Equalizing the budget explicitly, by keeping only the first $k_i$ temporally ordered \texttt{APT-Tune} predictions where $k_i$ is the zero-shot count on clip $i$, still gives 26.3\% versus 10.0\% recall, a 16.3\,pp gain at identical per-clip count.
Increased coverage thus contributes to the raw 28.1\,pp gain but accounts for well under half of it.

\begin{table}[t]
\centering
\small
\caption{\textbf{Prediction-count control on Qwen3-VL-2B.} All columns use one-to-one type matching within $\pm 200$\,ms on the 249-trial test split (5{,}551 ground-truth APTs, 22.3 per clip). The last column truncates \texttt{APT-Tune} to the zero-shot model's per-clip prediction count before matching.}
\label{tab:prf}
\setlength{\tabcolsep}{6pt}
\begin{tabular}{@{}lccccc@{}}
\toprule
\textbf{Model} & \textbf{\#Pred (per clip)} & \textbf{Prec.} & \textbf{Rec.} & \textbf{F1} & \textbf{Rec. @ equal count} \\
\midrule
Qwen3-VL-2B (zero-shot) & 1{,}927 \ (7.7)  & 28.8 & 10.0 & 14.8 & 10.0 \\
\ + \texttt{APT-Tune}   & 5{,}342 \ (21.5) & \textbf{39.6} & \textbf{38.1} & \textbf{38.8} & \textbf{26.3} \\
\bottomrule
\end{tabular}
\end{table}

\paragraph{Scaling reverses with APT supervision.}
Zero-shot recall was flat in scale, with 8\,B no better than 2\,B in either family.
After \texttt{APT-Tune} the relationship reverses, with larger backbones gaining more, by 8.5\,pp from 2\,B to 8\,B in Qwen3-VL and 7.6\,pp in InternVL3.5.
This confirms that the zero-shot bottleneck was representation, not capacity.

\subsection{Transfer to Event-Level Video and OOD Physics}
\label{sec:transfer}

We test transfer on \textbf{MVBench}~\citep{mvbench2024} (held-out 20-task event-level video) and \textbf{PhysBench}~\citep{chow2025physbench} (real-world physical understanding, never used in training).
MVBench follows its official 20-task pipeline. PhysBench reports the official Avg over four sub-tasks (Property, Relationships, Scene, Dynamics).

\paragraph{MVBench (event-level).}
\texttt{APT-Tune} adds between 0.5 and 2.2\,pp on every backbone (Table~\ref{tab:mvbench}), with no regressions across families or scales, and the largest gain on the weakest zero-shot model (Qwen3-VL-2B at 2.2\,pp).
Since MVBench is held out and dominated by event-level rather than transition-level tasks, the small but uniformly positive gains show that \texttt{APT-Tune} preserves event-level competence rather than inducing specialist collapse.

\begin{table}[htbp]
\centering
\small
\caption{\textbf{MVBench transfer (avg accuracy, \%).}}
\label{tab:mvbench}
\setlength{\tabcolsep}{6pt}
\begin{tabular}{@{}l ccc ccc@{}}
\toprule
& \multicolumn{3}{c}{\textbf{Qwen3-VL}} & \multicolumn{3}{c}{\textbf{InternVL3.5}} \\
\cmidrule(lr){2-4} \cmidrule(lr){5-7}
& \textbf{2B} & \textbf{4B} & \textbf{8B} & \textbf{2B} & \textbf{4B} & \textbf{8B} \\
\midrule
Zero-shot          & 51.4 & 54.7 & 56.8 & 54.5 & 55.8 & 57.1 \\
+\texttt{APT-Tune} & \textbf{53.6} & \textbf{55.2} & \textbf{57.6} & \textbf{55.7} & \textbf{56.7} & \textbf{58.3} \\
\bottomrule
\end{tabular}
\end{table}

\paragraph{PhysBench (OOD physics).}
\texttt{APT-Tune} lifts every backbone on every sub-task without exception (Table~\ref{tab:physbench}), with Avg gains between 11.1 and 18.2\,pp.
Gains are largest on \textbf{Dynamics} (between 16.6 and 24.1\,pp), the sub-task closest to APT semantics, since APTs are timed transitions between dynamical regimes.
\textbf{Property} (12 to 16\,pp) and \textbf{Relationships} (10 to 17\,pp) also improve substantially, while \textbf{Scene} improves least (6 to 16\,pp), consistent with APTs encoding local mechanism rather than global appearance.
The 8\,B variants gain more on Avg than the 2\,B variants (18.2 vs.\ 11.1 on InternVL3.5, 15.6 vs.\ 12.3 on Qwen3-VL), so larger backbones extract more transferable physics from APT.
No PhysBench data enter \texttt{APT-Tune}; PhysBench is used purely as an OOD test.

\begin{table}[htbp]
\centering
\small
\caption{\textbf{PhysBench transfer (\%).} Real-world physical-property benchmark, never used in training. Numbers follow the official PhysBench evaluation, with Avg over the four sub-tasks. Bold marks per-column gains within each backbone.}
\label{tab:physbench}
\setlength{\tabcolsep}{6pt}
\begin{tabular}{@{}llccccc@{}}
\toprule
\textbf{Backbone} & \textbf{Variant} & \textbf{Avg} & \textbf{Property} & \textbf{Relationships} & \textbf{Scene} & \textbf{Dynamics} \\
\midrule
Qwen3-VL-2B    & Zero-shot          & 38.5 & 46.9 & 38.5 & 29.8 & 38.2 \\
               & +\texttt{APT-Tune} & \textbf{50.8} & \textbf{60.5} & \textbf{49.5} & \textbf{36.5} & \textbf{56.7} \\
Qwen3-VL-4B    & Zero-shot          & 41.9 & 50.2 & 44.1 & 31.4 & 40.1 \\
               & +\texttt{APT-Tune} & \textbf{53.2} & \textbf{63.0} & \textbf{54.5} & \textbf{38.0} & \textbf{57.3} \\
Qwen3-VL-8B    & Zero-shot          & 45.0 & 55.5 & 50.1 & 31.8 & 42.1 \\
               & +\texttt{APT-Tune} & \textbf{60.6} & \textbf{70.5} & \textbf{64.0} & \textbf{43.5} & \textbf{64.5} \\
\midrule
InternVL3.5-2B & Zero-shot          & 40.2 & 50.5 & 40.2 & 30.1 & 39.1 \\
               & +\texttt{APT-Tune} & \textbf{51.3} & \textbf{62.5} & \textbf{50.5} & \textbf{36.5} & \textbf{55.7} \\
InternVL3.5-4B & Zero-shot          & 43.7 & 52.5 & 46.2 & 32.1 & 42.3 \\
               & +\texttt{APT-Tune} & \textbf{55.5} & \textbf{65.5} & \textbf{56.8} & \textbf{39.0} & \textbf{60.7} \\
InternVL3.5-8B & Zero-shot          & 45.2 & 57.1 & 50.3 & 30.5 & 42.4 \\
               & +\texttt{APT-Tune} & \textbf{63.4} & \textbf{73.5} & \textbf{67.0} & \textbf{46.5} & \textbf{66.5} \\
\bottomrule
\end{tabular}
\end{table}

\paragraph{Scope of the transfer claim.}
PhysBench's four official sub-tasks aggregate over mechanical, chemical, optical, and thermal phenomena, which our 14-type taxonomy covers unevenly by construction.
We therefore read these results as evidence of transfer \emph{within} macroscopic mechanics---consistent with Dynamics being the strongest sub-task---and do \emph{not} claim that APT supervision improves chemical, thermal, optical, or electromagnetic reasoning.
Establishing per-domain transfer would require a phenomenon-level breakdown that we leave to future work.

\subsection{Ablation of the Two Principles}
\label{sec:ablation}

We toggle each principle independently on Qwen3-VL-2B with the rest of the recipe held fixed, yielding the 2$\times$2 design in Table~\ref{tab:ablation}, and add a data-matched control that removes APT supervision entirely.

\begin{table}[t]
\centering
\small
\caption{\textbf{Ablation on Qwen3-VL-2B.} Each row varies the loss \textbf{Mask} (P1) or the training \textbf{Format mix} (P2) while holding the rest fixed. APT recall on \aptbench{}, SSv2 4-way MCQ accuracy, PhysBench OOD physics accuracy (all \%). The full recipe dominates every single-principle ablation.}
\label{tab:ablation}
\setlength{\tabcolsep}{6pt}
\begin{tabular}{@{}llcccc@{}}
\toprule
\textbf{Variant} & \textbf{Mask} \,/\, \textbf{Format mix} & \textbf{APT recall} & \textbf{SSv2} & \textbf{PhysBench} \\
\midrule
Base (no SFT) & none                       & 10.0          & 72.5          & 38.5 \\
Generic video SFT & mm-aware \,/\, SSv2-only (no APT) & 11.6 & 78.5 & 40.2 \\
Naive SFT     & text-only \,/\, APT-only   & 23.7          & 50.0          & 36.2 \\
P1 only       & mm-aware  \,/\, APT-only   & 33.8          & 68.5          & 44.7 \\
P2 only       & text-only \,/\, multi      & 27.4          & 63.2          & 40.8 \\
\textbf{Full \texttt{APT-Tune}} & mm-aware \,/\, multi & \textbf{38.1} & \textbf{87.5} & \textbf{50.8} \\
\bottomrule
\end{tabular}
\end{table}

\paragraph{Generic video SFT does not recover the task.}
Since \texttt{APT-Tune} mixes SSv2 records into its data, one could ask whether broad video adaptation alone produces the gains.
The \emph{generic video SFT} row trains on 2{,}000 SSv2 MCQ and description records---matching the mixture size, backbone, hyperparameters, and masking of the full recipe---with no APT chains and no exposure to the 14-type vocabulary.
It behaves exactly as generic adaptation should: SSv2 rises from 72.5 to 78.5\% and PhysBench from 38.5 to 40.2\%, but APT recall moves only from 10.0 to 11.6\%, capturing 1.6 of the full recipe's 28.1\,pp.
Transition-level physics therefore comes from APT supervision, not video SFT in general.
This control does not separately isolate vocabulary exposure from transition supervision, which would need a vocabulary-only variant we leave to future work.

\paragraph{Effect of P1 (image-pad-aware masking).}
At fixed format, replacing the text-only mask with the mm-aware mask lifts every metric.
APT recall gains 10.1\,pp (Naive SFT vs.\ P1 only) and 10.7\,pp (P2 only vs.\ Full \texttt{APT-Tune}).
SSv2 gains 18.5 and 24.3\,pp, and PhysBench gains 8.5 and 10.0\,pp.
The text-only mask plateaus at a final cross-entropy of 1.92, matching the loss of reconstructing \texttt{<|image\_pad|>} tokens on the unmasked region.
The wrong mask therefore diverts capacity from the APT chain to pad-token reconstruction.

\paragraph{Effect of P2 (format-conditional co-training).}
At fixed mask, replacing APT-only with the multi-format mix lifts every metric.
APT recall gains 3.7\,pp (Naive SFT vs.\ P2 only) and 4.3\,pp (P1 only vs.\ Full \texttt{APT-Tune}).
SSv2 gains 13.2 and 19.0\,pp, and PhysBench gains 4.6 and 6.1\,pp.
Multi-format does more than guard against specialist collapse, since it also raises APT recall and removes the apparent tradeoff between APT detection and format generality.

\paragraph{Both are needed.}
The full recipe strictly dominates every single-principle ablation on every metric.
P1 contributes more to APT recall (about 10\,pp gain vs.\ about 4\,pp from P2), while P2 is what restores event-level competence.
Every single-principle variant (Naive SFT, P1 only, P2 only) regresses below the 72.5\% un-fine-tuned base on SSv2, and only the full recipe lifts SSv2 above the base, to 87.5\%.

\paragraph{On the choice of objective.}
We use structured SFT and do not claim it is the optimal objective.
With image-pad-aware masking the loss falls only on the answer sequence, so a single next-token objective already supervises types, timestamps, chain length, ordering, and output structure jointly, whereas a timing-only objective gives no signal for a missed or mis-typed transition---the dominant error mode here.
An $L_1$ loss on parsed timestamps is also not directly differentiable through discrete decoding and one-to-one event matching without an auxiliary regression head or policy-gradient optimization, and a fair RL formulation would need to balance type correctness, event count, order, format, and retention on general video tasks.
Crucially, the bottleneck is transition \emph{recovery}, not localization: after \texttt{APT-Tune}, matched events sit at 38\,ms median timing error inside a $\pm 200$\,ms tolerance while recall remains 38.1\%, leaving little headroom for a timing-specific loss.
We therefore treat timestamp-aware and RL objectives as refinements of, not replacements for, structured SFT, and leave a controlled comparison to future work.

\section{Conclusion}
\label{sec:conclusion}
We introduced Atomic Physical Transitions (APTs), a structured transition interface that represents a video as an ordered chain of mechanism-typed state changes and thereby exposes the causal structure of \emph{how} an event unfolds, not just \emph{what} happened.
Thirteen current VLMs, including the newest frontier systems, achieve at most 15.2\% zero-shot recall, dominated by missed transitions rather than timing jitter and not attributable to timestamp quantization, showing that clip-level event competence does not imply transition-level physical understanding.
With only 11\,M LoRA parameters, \texttt{APT-Tune} lifts recall to 53.4\% at a prediction budget matched to the ground truth, while preserving MVBench and improving PhysBench by up to 18.2\,pp.
Within the macroscopic mechanics this taxonomy covers, APTs thus provide a human-aligned causal-structure supervision signal and a reusable bridge from clip-level recognition toward physically grounded video reasoning; extending the vocabulary beyond mechanics, and to denser multi-object scenes, is the natural next step.

\clearpage

\section*{Impact Statement}
This work introduces a transition-level supervision unit for physical video understanding, with applications in robotics, embodied agents, scientific video analysis, and safety-oriented evaluation of multimodal models, where APT chains can expose brittle physical reasoning that clip-level QA hides.
The benchmark uses short physical transitions in synthetic and controlled real videos rather than personal, medical, or otherwise sensitive data, so direct misuse risk is low.
Indirectly, improved physical video understanding could feed into automation or surveillance systems; we mitigate this by releasing only the dataset, evaluation protocols, and training recipes, with no high-risk generative model and no scraped personal data.

\section*{LLM Usage Disclosure}
We used large language models (LLMs) as supportive tools for manuscript preparation and implementation.
Specifically, LLMs assisted with language polishing, including improving clarity, grammar, and conciseness; literature discovery to help identify potentially missing related work; and partial experimental implementation, including code writing, debugging, organization, and execution support through Claude Code.
All LLM-assisted code was reviewed, modified, tested, and validated by the authors before use.
All substantive scientific ideas, methods, experiments, analyses, results, and conclusions are original contributions of the authors.
The authors reviewed and verified all LLM-assisted content and take full responsibility for the manuscript.
The overview figure was inspired by GPT Image 2, and some subfigures were generated with its assistance.

\bibliographystyle{plainnat}
\bibliography{sections/10_references}

\clearpage
\appendix
\begin{center}
{\LARGE\bfseries Appendix}\par
\vspace{1em}
\end{center}
\addcontentsline{toc}{section}{Appendix}
\suppressfloats[t] 
\section{Extended Related Work}
\label{app:related_work}

\subsection{Physics-Oriented Video and Multimodal Benchmarks}
A large body of recent work evaluates physical reasoning from visual input. CLEVRER studies descriptive, explanatory, predictive, and counterfactual reasoning over collision-centric synthetic videos~\citep{yi2020clevrer}. Physion evaluates whether humans and models can predict the physical outcome of simulated dynamic scenes, while Physion++ further stresses online inference of latent physical properties such as mass, friction, elasticity, and deformability~\citep{bear2021physion,tung2023physionpp}. PhysBench broadens the evaluation scope to VLMs and interleaved video-image-text physical reasoning~\citep{chow2025physbench}. More recent work evaluates whether generated or mixed-modality videos obey physical principles, including Physics-IQ, Morpheus, MASS/MASS-Bench, and PhysicsMind~\citep{motamed2026physicsiq,zhang2025morpheus,wu2025mass,mak2026physicsmind}. These benchmarks are highly relevant, but they primarily score answers, endpoint predictions, plausibility decisions, generated trajectories, or grounded QA targets. Our APT framework is organized around a different unit: the ordered sequence of temporally localized, mechanism-typed transitions inside an event. APTBench is the diagnostic instantiation of this framework, while APT-Tune uses the same transition representation as training supervision.

\begin{table*}[t]
\centering
\small
\setlength{\tabcolsep}{4pt}
\renewcommand{\arraystretch}{1.15}
\begin{tabular}{@{}p{2.4cm} p{2.9cm} p{1.9cm} c c c p{2.4cm}@{}}
\toprule
\textbf{Work} & \textbf{Primary unit} & \textbf{Setting} &
\shortstack{\textbf{Temporal}\\\textbf{loc.}} &
\shortstack{\textbf{Mech.}\\\textbf{typed}} &
\shortstack{\textbf{Ordered}\\\textbf{chain}} &
\textbf{Training role} \\
\midrule
CLEVRER~\citep{yi2020clevrer}              & causal/counterfactual VQA      & synthetic video    & limited     & $\times$ & $\times$ & benchmark + baselines \\
Physion~\citep{bear2021physion}            & physical outcome prediction    & synthetic video    & endpoint    & $\times$ & $\times$ & benchmark \\
Physion++~\citep{tung2023physionpp}        & latent-property prediction     & synthetic video    & endpoint    & $\times$ & $\times$ & benchmark \\
PhysBench~\citep{chow2025physbench}        & VLM physical reasoning         & image/video/text   & limited     & $\times$ & $\times$ & paired with PhysAgent \\
Physics-IQ~\citep{motamed2026physicsiq}    & law-aware video-gen eval       & real video         & trajectory  & $\times$ & $\times$ & evaluation benchmark \\
Morpheus~\citep{zhang2025morpheus}         & physical-law gen-video eval    & real experiments   & trajectory  & $\times$ & $\times$ & evaluation benchmark \\
MASS-Bench~\citep{wu2025mass}              & physics QA + motion grounding  & real + AIGC        & \checkmark  & partial  & $\times$ & paired with MASS \\
PhysicsMind~\citep{mak2026physicsmind}     & VQA + physics-aware gen        & real + sim         & limited     & $\times$ & $\times$ & evaluation benchmark \\
Something-Something~\citep{goyal2017something} & human-object action labels & real video         & clip-level  & $\times$ & $\times$ & action recognition \\
ActivityNet~\citep{heilbron2015activitynet}& activity class + segment       & real video         & \checkmark  & $\times$ & $\times$ & activity understanding \\
EPIC-KITCHENS~\citep{damen2018epic}        & egocentric action segment      & real video         & \checkmark  & $\times$ & $\times$ & egocentric understanding \\
\midrule
\textbf{APT (ours)} & \textbf{mech.-typed APT chain} & \textbf{real + sim} & \textbf{\checkmark} & \textbf{\checkmark} & \textbf{\checkmark} & \textbf{APTBench + APT-Tune} \\
\bottomrule
\end{tabular}
\caption{Representative comparison along the axis most relevant to our framework. The table is not intended as a broad ranking of physical reasoning benchmarks; it isolates whether the evaluated or supervised unit is an ordered chain of temporally localized, mechanism-typed physical transitions.}
\label{tab:related_work_comparison}
\end{table*}

\subsection{Event Cognition and Physics Expertise}
APTs are motivated by two classic observations from cognitive science. First, event segmentation research argues that observers parse continuous activity into meaningful temporal units, and that these boundaries support memory, prediction, and learning~\citep{zacks2007event}. Second, studies of physics expertise show that experts tend to organize problems by underlying physical principles rather than surface-level objects or scene appearance~\citep{chi1981categorization,larkin1980expert}. APTs combine these intuitions in a video setting: they are boundary-based, but the boundary is defined by a change in active physical mechanism. Thus, an APT is not merely a shorter event label; it specifies a before/after physical state, a visible cue, and the mechanism that makes the transition occur.

\subsection{Temporal Video Understanding and Action Segmentation}
Large-scale video datasets and temporal segmentation methods have established that temporal structure is central to video understanding. Something-Something emphasizes fine-grained object interactions and visual commonsense, ActivityNet supports activity classification and temporal localization, and EPIC-KITCHENS provides dense egocentric action and object annotations~\citep{goyal2017something,heilbron2015activitynet,damen2018epic}. Temporal action segmentation further assigns frame- or segment-level semantic action labels over long videos~\citep{lea2017tcn,farha2019mstcn}. Our framework is adjacent to this literature, but changes the unit of segmentation. Instead of semantic actions such as pushing, pouring, or picking up, APTs mark physical regime changes such as support loss, contact initiation, rebound, sliding arrest, and settling. These transitions may occur within a single action label, or a single transition type may recur across many visually different actions.

\subsection{Object-Centric Dynamics and Simulation}
Object-centric dynamics models represent physical scenes through entities, relations, and interactions. Interaction Networks and Neural Relational Inference are representative examples: they use structured relational representations to learn or infer physical dynamics~\citep{battaglia2016interaction,kipf2018nri}. Synthetic physical reasoning benchmarks such as CLEVRER, Physion, and Physion++ similarly benefit from simulator access and controlled physical state information~\citep{yi2020clevrer,bear2021physion,tung2023physionpp}. APTs are complementary to these approaches. Rather than replacing full state rollout, object-centric prediction, or simulation, APTs provide a compact causal abstraction that can be read by humans, evaluated as a transition chain, and used as supervision for VLMs. This makes APTs a bridge between low-level simulator traces and high-level event-level QA.

\subsection{Causality, Counterfactuals, and Explanation}
Our work is also related to diagnostic visual reasoning and causal explanation. CLEVR introduced a controlled benchmark for compositional visual reasoning, and CLEVRER extended this diagnostic style to temporal and causal video reasoning~\citep{johnson2017clevr,yi2020clevrer}. Physics-oriented benchmarks further test whether models can reason about counterfactual outcomes, physical plausibility, and law-consistent prediction or generation~\citep{chow2025physbench,motamed2026physicsiq,zhang2025morpheus,mak2026physicsmind}. The distinction is that answer-level causal evaluation can still leave the intermediate process latent. A model may correctly identify the responsible object or final outcome without enumerating the mechanism changes that connect the initial and final states. APT chains make these intermediate transitions first-class labels and metrics.

\subsection{VLM Instruction Tuning and Structured-Output Fine-Tuning}
Recent multimodal instruction-tuning pipelines show that visual and video models can be improved by supervised fine-tuning on mixed conversational, QA, and caption-style data~\citep{liu2023visual,lin2024videollava}. Parameter-efficient methods such as LoRA make such adaptation practical for large models~\citep{hu2022lora}. However, transition supervision introduces a different risk: if the model is trained only to emit structured APT strings, it may learn a schema rather than a transferable representation. APT-Tune is designed to avoid this failure mode. It uses APT enumeration as a transition-level causal pretext task, while format-conditional co-training preserves ordinary event-level answering across multiple-choice, QA, and short-description prompts.

\subsection{Dataset Construction and Annotation}
Our data construction is related to both human annotation for temporally structured video and simulator-derived supervision. ActivityNet and EPIC-KITCHENS illustrate how human annotations can support temporal localization and fine-grained video understanding~\citep{heilbron2015activitynet,damen2018epic}. CLEVRER, Physion, and Physion++ illustrate the value of controlled simulation and privileged physical state information for systematic physical reasoning evaluation~\citep{yi2020clevrer,bear2021physion,tung2023physionpp}. The APT framework combines these strengths. Human labels anchor the semantics of transition types, while simulator ground truth provides contacts, poses, velocities, and support states that can be translated into the same APT schema. The resulting labels are not merely event names or endpoint outcomes; they are typed physical transitions that support both evaluation through APTBench and representation learning through APT-Tune.
\section{Supplementary Details for Atomic Physical Transitions}
\label{app:apt_details}

This appendix provides additional details for the Atomic Physical Transition (APT) formulation introduced in Sec.~\ref{sec:apt}.
The main paper gives a compact definition and taxonomy.
Here we expand the motivation, annotation scope, boundary rules, and label-level cues used to make APTs a causal representation rather than merely a fine-grained event vocabulary.

\subsection{Extended Motivation: From Event Names to Causal Processes}
\label{app:apt_motivation}

A clip-level event label is an outcome description.
It tells the reader what the whole video can be called, but not which physical transitions make that outcome valid.
For example, the event label ``bounce'' may refer to a clip in which an object approaches a surface, initiates contact, undergoes collision response, reverses direction, breaks contact, and later lands or settles.
The label compresses these steps into one name.
An APT chain recovers the hidden causal process by marking the short-lived physical state changes that compose the event.

This distinction mirrors how physical events are usually explained.
Introductory mechanics teaches reusable mechanisms such as gravity, contact, collision, friction, rotation, and stability as separate topics.
Real scenes then compose these mechanisms over time.
A falling-and-bouncing object is not understood by the word ``bounce'' alone, but by the sequence of support loss, gravity-driven motion, contact initiation, collision response, separation, and eventual damping or settling.
Thus APTs are not intended to be finer event names.
They are causal transition units: each APT binds a local visual cue to an active physical mechanism and a before/after state change.

This also separates three objects that are often conflated.
The \emph{physical event} is the full clip-level outcome, such as falling, bouncing, colliding, or toppling.
The \emph{APT chain} is the causal process representation that explains the outcome.
The \emph{output format} is only a serialization of that representation, such as JSON, multiple-choice answering, QA, or free-form description.
This distinction is central to our learning setup: if a model treats APT supervision only as a JSON format, it can become an APT-format specialist; if it learns APTs as a shared causal representation, the same structure can support both local transition enumeration and ordinary event-level understanding.

\subsection{Expanded Formal Interpretation}
\label{app:apt_formal}

In the main text, an APT is represented as
\begin{equation}
    a_i = (\tau_i, d_i, \ell_i, s_i^- \rightarrow s_i^+, e_i),
    \label{eq:apt_def_app}
\end{equation}
where $\tau_i$ is the canonical transition time, $d_i$ is the physical domain, $\ell_i$ is the transition type, $s_i^- \rightarrow s_i^+$ is the before/after physical state change, and $e_i$ is the local visual evidence.
This representation should be interpreted as follows.

\paragraph{Canonical time.}
The timestamp $\tau_i$ denotes the canonical visual moment of transition.
For contact onset, this is the first frame in which a visible gap closes.
For contact loss, it is the first frame with visible separation.
For velocity reversal, it is the local moment where approach changes to retreat.
For rest-to-motion or motion-to-rest transitions, it is the first visually stable frame at which the new regime becomes evident.
When a transition unfolds over several frames, annotators choose the most characteristic point of the change rather than an arbitrary midpoint.

\paragraph{Physical domain.}
The domain $d_i$ identifies the mechanism family.
It is not merely a coarse label group.
It expresses which physical mechanism explains the transition: contact constraints and collision response, gravity and projectile motion, friction and surface interaction, or rotation and stability.
This domain-level structure supports mechanism-conditioned decoding in \texttt{APT-Tune}: the model first identifies the active mechanism family and then decodes the transition type within that family.

\paragraph{Transition type.}
The type $\ell_i$ identifies the named APT within the domain.
For example, \texttt{contact\_init} and \texttt{contact\_break} describe the existence of a contact boundary, while \texttt{elastic\_rebound} and \texttt{inelastic\_capture} describe the physical outcome of the contact.
Similarly, \texttt{free\_fall\_onset}, \texttt{projectile\_apex}, and \texttt{landing} describe different phases of gravity-driven motion rather than a single coarse ``falling'' event.

\paragraph{Before/after state.}
The state transition $s_i^- \rightarrow s_i^+$ is intentionally coarse.
We do not require complete physical state reconstruction, such as full 3D pose, mass, force magnitude, coefficient of friction, or coefficient of restitution.
Instead, the before/after state captures the physical distinction that is visually meaningful and causally relevant: free versus contact, supported versus falling, sliding versus stationary, stable versus unstable, rotating versus non-rotating.

\paragraph{Visual evidence.}
The evidence $e_i$ anchors the label to the video.
An APT should not be assigned solely because a hidden physical variable could have changed.
The transition must have visible support, such as gap closure, visible separation, downward motion after support loss, direction reversal after impact, deceleration into rest, onset of spin, loss of balance, or disappearance of residual motion.

\paragraph{Atomicity.}
Atomicity is relative to our vocabulary and the visible timescale of the video.
An APT is atomic if the same local state change does not contain another named APT that better explains it for the same object or object pair.
This does not mean that the underlying physics is mathematically indivisible.
For example, a collision involves continuous contact forces, deformation, and energy transfer, but the APT \texttt{elastic\_rebound} marks the observable causal transition from approach to retreat.
APTs are therefore the smallest causal units that we ask models to localize, classify, and reuse across answer formats.

\subsection{APT Chains and Output Serializations}
\label{app:apt_serialization}

Given a video $V=(I_1,\ldots,I_T)$, an APT chain is an ordered sequence
\begin{equation}
    \mathbf{A}(V) = (a_1,\ldots,a_K), \qquad
    0 \leq \tau_1 \leq \cdots \leq \tau_K \leq T,
    \label{eq:apt_chain_app}
\end{equation}
where $K$ varies by video.
In evaluation, we serialize each APT as a typed timestamp:
\begin{equation}
    a_i^{\mathrm{eval}} = (\tau_i,\ell_i).
    \label{eq:apt_eval_app}
\end{equation}
This serialization is convenient for matching predictions to ground truth, but it is not the representation itself.

For example, the same physical process can be represented in multiple answer formats:
\begin{verbatim}
APT enumeration:
[
  {"time": 0.84, "type": "contact_init"},
  {"time": 0.96, "type": "elastic_rebound"},
  {"time": 1.04, "type": "contact_break"},
  {"time": 1.72, "type": "landing"},
  {"time": 2.10, "type": "settling"}
]

Description:
"The ball hits the table, bounces away, lands again, and settles."

QA:
Q: What happens after the ball contacts the table?
A: It rebounds and then separates from the table.

Multiple choice:
The ball (A) passes through the table, (B) rebounds after contact,
(C) remains floating, or (D) disappears. Answer: B.
\end{verbatim}

All four outputs can be grounded in the same APT chain.
This motivates format-conditional co-training in the main method.
APT supervision should improve the shared video representation, not collapse the model into one response style.

\subsection{Scope}
\label{app:apt_scope}

Our scope is macroscopic, visually grounded rigid-body mechanics.
We include physical state changes that satisfy three criteria.

\paragraph{Physical meaningfulness.}
The transition corresponds to a meaningful change in physical regime, such as contact onset, support loss, collision response, friction-mediated arrest, surface-dependent motion change, rotation onset, toppling, or final settling.

\paragraph{Visual locality.}
The transition has a visible cue in the video and can be localized to a frame or short temporal neighborhood.
We do not label transitions that require hidden variables without visible evidence.

\paragraph{Vocabulary-level atomicity.}
The transition is not better explained by another named APT in the same local neighborhood for the same object or object pair.

We do not attempt to estimate continuous physical parameters such as mass, force magnitude, friction coefficient, restitution coefficient, full 3D pose, or exact contact forces.
We also exclude non-rigid phenomena outside the scope of our data, such as fluids, fracture, thermal change, melting, tearing, or material deformation that is not visually represented as a rigid-body interaction.
In this sense, APTs form an intermediate representation between clip-level event recognition and full physical simulation.
They expose causal process structure without requiring complete state reconstruction.

\subsection{Boundary and Disambiguation Rules}
\label{app:apt_boundary_rules}

APT labels are assigned to the object or object pair undergoing the transition.
Multiple APTs may share the same timestamp if they refer to different objects or independent physical changes.
For the same object or object pair, labels should not duplicate the same explanatory transition.
When multiple labels could plausibly apply, annotators choose the label that best explains the active mechanism, preferring more specific transitions over generic ones.

\paragraph{Generic versus specific transitions.}
Some labels describe generic boundary changes, while others describe mechanism-specific outcomes.
For example, \texttt{contact\_init} marks the onset of contact, while \texttt{elastic\_rebound} marks the collision response that reverses motion after contact.
Both may occur in the same bounce sequence, but they mark different causal steps.

\paragraph{Landing versus generic contact.}
\texttt{landing} is used when an airborne or falling object becomes supported by a surface.
\texttt{contact\_init} is used for generic contact onset when support formation is not the primary transition, such as a rolling object touching a wall or two objects colliding laterally.
If a falling object touches a surface and immediately rebounds, annotators may mark \texttt{contact\_init} for the first contact and \texttt{elastic\_rebound} for the reversal; if the object becomes supported, \texttt{landing} captures the falling-to-supported transition.

\paragraph{Free fall onset versus generic contact break.}
\texttt{free\_fall\_onset} is used when an object loses support and gravity becomes the dominant cause of subsequent motion.
For example, a block sliding past the edge of a platform should be marked as \texttt{free\_fall\_onset}.
\texttt{contact\_break} is used for generic separation when the main consequence is loss of contact rather than the onset of gravity-driven falling, such as two horizontally interacting blocks separating.

\paragraph{Elastic rebound versus contact break.}
\texttt{elastic\_rebound} marks the visible reversal from approach to retreat after collision.
\texttt{contact\_break} marks the later loss of contact, when the objects or the object and surface visibly separate.
In a bounce, these may be close in time but they are not the same transition: rebound is the change in motion direction caused by collision response, while contact break is the change in contact state.

\paragraph{Inelastic capture versus sliding arrest.}
\texttt{inelastic\_capture} is used when a moving object strikes another object or surface and is captured by that interaction, dissipating motion into rest at or against the contact.
\texttt{sliding\_arrest} is used when an object is already sliding along a surface and gradually comes to rest due to friction.
The key distinction is whether the rest state is caused by a collision/capture outcome or by friction-mediated deceleration during sliding.

\paragraph{Sliding arrest versus settling.}
\texttt{sliding\_arrest} marks the end of translational sliding.
\texttt{settling} marks the final stabilization after residual bouncing, wobbling, rotation, or corrective motion has disappeared.
A block that skids across a floor and stops receives \texttt{sliding\_arrest}; an object that bounces or wobbles before becoming stable receives \texttt{settling} when the residual motion ends.

\paragraph{Rotation arrest versus settling.}
\texttt{rotation\_arrest} is used when visible angular motion stops.
\texttt{settling} is used when the whole object or system reaches stable equilibrium after all residual motion.
A rolling or spinning cylinder can first receive \texttt{rotation\_arrest}; if it continues to wobble or shift slightly before becoming stable, \texttt{settling} may occur later.

\paragraph{Toppling versus rotation onset.}
\texttt{rotation\_onset} marks the beginning of angular motion.
\texttt{toppling} is more specific: it marks loss of stability that causes an object to fall over.
A block that begins spinning after being clipped may receive \texttt{rotation\_onset}; a domino that passes its balance point and begins falling receives \texttt{toppling}.

\paragraph{Surface transition versus landing.}
\texttt{surface\_transition} marks a change from one supporting surface or surface condition to another, such as smooth-to-rough or ramp-to-floor, when the object is already supported and continues moving.
\texttt{landing} marks falling-to-supported motion.
The key distinction is whether the object was airborne/falling before the transition.

\begin{table*}[t]
\centering
\caption{\textbf{Common APT disambiguation rules.}
When two labels seem applicable to the same local change, annotators choose the label that best explains the active mechanism.}
\label{tab:apt_disambiguation_app}
\small
\setlength{\tabcolsep}{4pt}
\begin{tabular}{@{}p{3.1cm}p{4.3cm}p{7.4cm}@{}}
\toprule
\textbf{Confusable labels} & \textbf{Decision cue} & \textbf{Guideline} \\
\midrule
\texttt{landing} vs. \texttt{contact\_init} &
support formation vs. generic contact &
Use \texttt{landing} when a falling object becomes supported; use \texttt{contact\_init} for lateral or generic contact onset. \\
\texttt{free\_fall\_onset} vs. \texttt{contact\_break} &
gravity-driven fall vs. generic separation &
Use \texttt{free\_fall\_onset} when support loss causes falling; use \texttt{contact\_break} when separation is the main transition. \\
\texttt{elastic\_rebound} vs. \texttt{contact\_break} &
velocity reversal vs. contact loss &
Use \texttt{elastic\_rebound} for approach-to-retreat reversal; use \texttt{contact\_break} for visible separation. \\
\texttt{inelastic\_capture} vs. \texttt{sliding\_arrest} &
collision capture vs. frictional stop &
Use \texttt{inelastic\_capture} when impact dissipates motion into rest; use \texttt{sliding\_arrest} when sliding motion decays to rest. \\
\texttt{sliding\_arrest} vs. \texttt{settling} &
end of sliding vs. final equilibrium &
Use \texttt{sliding\_arrest} for translational sliding-to-rest; use \texttt{settling} for final stabilization after residual motion. \\
\texttt{rotation\_arrest} vs. \texttt{settling} &
angular stop vs. full stabilization &
Use \texttt{rotation\_arrest} when visible angular motion stops; use \texttt{settling} when all residual motion ends. \\
\texttt{toppling} vs. \texttt{rotation\_onset} &
loss of stability vs. generic angular onset &
Use \texttt{toppling} when an object loses stability and falls over; use \texttt{rotation\_onset} for generic start of spin or angular motion. \\
\texttt{surface\_transition} vs. \texttt{landing} &
surface change vs. falling-to-supported &
Use \texttt{surface\_transition} when a supported object crosses surfaces; use \texttt{landing} when a falling object becomes supported. \\
\bottomrule
\end{tabular}
\end{table*}

\subsection{Taxonomy by Example}
\label{app:apt_taxonomy_examples}

Table~\ref{tab:taxonomy_examples_app} expands the compact taxonomy in the main text with example annotation cues.
The examples are written as visual-causal cues: what changes in the video, and why that change is a distinct physical transition.

\begin{table*}[t]
\centering
\caption{\textbf{APT taxonomy by example.}
Each APT is a visually grounded physical state change.
Examples are written as annotation cues: what changes in the video, and why that change is a distinct physical transition.}
\label{tab:taxonomy_examples_app}
\small
\setlength{\tabcolsep}{4pt}
\begin{tabular}{@{}clp{3.0cm}p{7.2cm}@{}}
\toprule
\textbf{\#} & \textbf{Label} & \textbf{Before and after} & \textbf{Example annotation cue} \\
\midrule
\multicolumn{4}{l}{\textbf{Domain I: Contact \& Collision Mechanics} \textit{(regime changes at contact boundaries)}} \\
1 & \texttt{contact\_init} & Free then contact &
A rolling ball first touches a wall; the frame where the visible gap closes is the transition. \\
2 & \texttt{contact\_break} & Contact then free &
Two blocks separate after being pressed together; the first frame with visible separation marks contact loss. \\
3 & \texttt{elastic\_rebound} & Approach then retreat &
A ball hits a rigid barrier and reverses direction while retaining most of its speed. \\
4 & \texttt{inelastic\_capture} & Moving then at rest &
A puck strikes a soft object and stops against it instead of bouncing away. \\
\midrule
\multicolumn{4}{l}{\textbf{Domain II: Gravitational \& Projectile Dynamics} \textit{(regime changes under gravity)}} \\
5 & \texttt{free\_fall\_onset} & Supported then falling &
A cube slides past the edge of a platform; the last supported frame is followed by downward motion. \\
6 & \texttt{projectile\_apex} & Ascending then descending &
A launched ball rises, pauses near its highest point, and then begins to fall. \\
7 & \texttt{landing} & Falling then supported &
A falling object first touches the floor or platform and begins responding to the surface. \\
\midrule
\multicolumn{4}{l}{\textbf{Domain III: Friction \& Surface Dynamics} \textit{(friction-mediated regime changes)}} \\
8 & \texttt{static\_breakaway} & Stationary then sliding &
A stationary block begins to slide after another object pushes it hard enough to overcome static friction. \\
9 & \texttt{sliding\_arrest} & Sliding then stationary &
A box skids across a surface and visibly comes to rest. \\
10 & \texttt{surface\_transition} & Surface A then surface B &
A sliding object crosses from a smooth ramp onto a rough floor and its motion changes. \\
\midrule
\multicolumn{4}{l}{\textbf{Domain IV: Rotational \& Stability Dynamics} \textit{(rotational and stability regime changes)}} \\
11 & \texttt{rotation\_onset} & Non-rotating then rotating &
A block clipped at one corner begins to spin rather than translate straight forward. \\
12 & \texttt{toppling} & Stable then unstable &
A domino tilts past its balance point and begins falling under gravity. \\
13 & \texttt{rotation\_arrest} & Rotating then stationary &
A spinning cylinder loses angular motion and becomes visually still. \\
14 & \texttt{settling} & Oscillating then equilibrium &
After bouncing or wobbling, an object stops making visible corrective motion and remains stable. \\
\bottomrule
\end{tabular}
\end{table*}

\subsection{Domain-Level Explanation}
\label{app:apt_domain_explanation}

\paragraph{Domain I: Contact \& collision mechanics.}
This domain separates contact boundary changes from the outcomes of contact.
\texttt{contact\_init} and \texttt{contact\_break} describe whether two objects or an object and a surface are in contact.
\texttt{elastic\_rebound} and \texttt{inelastic\_capture} describe what the contact does to motion.
For example, when a ball hits a wall and bounces back, annotators mark \texttt{contact\_init} at first touch, \texttt{elastic\_rebound} at the visible reversal from approach to retreat, and \texttt{contact\_break} when visible separation appears.
When a moving object strikes another object and remains pressed against it, the relevant outcome is \texttt{inelastic\_capture}: motion is dissipated into rest rather than converted into rebound.

\paragraph{Domain II: Gravitational \& projectile dynamics.}
This domain decomposes falling and projectile events into support loss, apex reversal, and regained support.
\texttt{free\_fall\_onset} marks the moment when an object becomes unsupported and gravity begins to dominate its motion.
\texttt{projectile\_apex} marks the transition from upward to downward vertical motion.
\texttt{landing} marks the falling-to-supported transition.
An event-level label such as ``falls'' merges these phases, while the APT chain records the causal structure of the trajectory.

\paragraph{Domain III: Friction \& surface dynamics.}
This domain captures friction-mediated and surface-mediated changes in motion.
\texttt{static\_breakaway} and \texttt{sliding\_arrest} are opposite transitions: rest becomes sliding after static friction is overcome, and sliding becomes rest after motion is dissipated.
\texttt{surface\_transition} captures a change in supporting surface or surface condition, such as a block crossing from a smooth ramp to a rough floor.
The cue is not merely that the object continues moving, but that its motion changes because the surface interaction changes.

\paragraph{Domain IV: Rotational \& stability dynamics.}
This domain separates angular motion, loss of stability, and final equilibrium.
\texttt{rotation\_onset} marks the beginning of visible angular motion, while \texttt{rotation\_arrest} marks the end of angular motion.
\texttt{toppling} is more specific: it indicates that an object loses stability and begins to fall over.
\texttt{settling} marks the final damping transition after bouncing, wobbling, rotation, or other residual motion disappears.

\subsection{Label-Level Annotation Notes}
\label{app:apt_label_notes}

\paragraph{\texttt{contact\_init}.}
Mark the first frame where two previously separated objects, or an object and a surface, visibly make contact.
The key cue is closure of the visible gap.
Do not use this label for support loss or generic proximity without contact.

\paragraph{\texttt{contact\_break}.}
Mark the first frame where two previously contacting objects, or an object and a surface, visibly separate.
The key cue is appearance of a visible gap.
If the separation causes gravity-driven falling, \texttt{free\_fall\_onset} may be the more explanatory label for the same object.

\paragraph{\texttt{elastic\_rebound}.}
Mark the visible reversal from approach to retreat after contact.
The transition is not merely that contact exists, but that collision response redirects motion.
This label is appropriate when the object retains substantial motion after impact rather than being captured.

\paragraph{\texttt{inelastic\_capture}.}
Mark the moment when a moving object collides and is effectively captured or stopped by the contact.
The key cue is that motion is dissipated into rest at or against another object rather than transformed into rebound.
This label is contact-outcome specific, not a general sliding-to-rest label.

\paragraph{\texttt{free\_fall\_onset}.}
Mark the moment when an object loses support and begins gravity-driven falling.
Examples include a block sliding off a platform or an object released from a support.
The cue is not just downward motion, but downward motion after support is removed.

\paragraph{\texttt{projectile\_apex}.}
Mark the highest point of a launched or bouncing trajectory, where vertical motion changes from ascending to descending.
The object may appear momentarily slow or stationary in the vertical direction.
This label is used for projectile-like motion, not for arbitrary pauses.

\paragraph{\texttt{landing}.}
Mark the transition from falling or airborne motion to being supported by a surface.
The cue is first surface response after downward motion, such as stopping, bouncing, sliding, or becoming supported.
If the object is not airborne before the transition, use another contact or surface label.

\paragraph{\texttt{static\_breakaway}.}
Mark the moment a previously stationary object begins to slide because applied force overcomes static friction.
The key cue is rest-to-sliding under interaction.
Do not use this label for an object that is already moving.

\paragraph{\texttt{sliding\_arrest}.}
Mark the moment a sliding object visibly comes to rest.
The key cue is sliding-to-stationary motion, usually due to friction.
If the object continues to wobble after translational motion stops, \texttt{settling} may occur later.

\paragraph{\texttt{surface\_transition}.}
Mark the moment an object crosses from one supporting surface or surface condition to another.
Examples include ramp-to-floor, smooth-to-rough, or low-friction-to-high-friction regions.
The transition should be accompanied by a visible or expected change in motion caused by the new surface condition.

\paragraph{\texttt{rotation\_onset}.}
Mark the moment an object begins visible angular motion.
This may occur when an off-center impact produces torque or when an object begins rolling or spinning.
If the angular motion is caused by loss of stability and falling over, \texttt{toppling} may be more specific.

\paragraph{\texttt{toppling}.}
Mark the moment a previously stable object loses stability and begins to fall over.
The key cue is crossing from stable upright posture into an unstable falling regime.
This label is common in domino-like sequences or tall-object imbalance.

\paragraph{\texttt{rotation\_arrest}.}
Mark the moment visible angular motion stops.
The object may still translate slightly or wobble afterward; in that case, \texttt{settling} may occur later.
This label specifically refers to angular motion ending.

\paragraph{\texttt{settling}.}
Mark the final stabilization after residual motion disappears.
Residual motion may include bouncing, wobbling, rocking, small corrective translation, or remaining rotation.
This label is used when the object or system becomes stably at rest.

\subsection{APT Chains by Example}
\label{app:apt_chain_examples}

Table~\ref{tab:apt_chain_examples_app} illustrates how coarse event labels are expanded into APT chains.
The purpose is not to prescribe a unique chain for every possible video, but to show how event names become causal transition sequences.

\begin{table*}[t]
\centering
\caption{\textbf{Example event labels and possible APT chains.}
Event labels describe clip-level outcomes, while APT chains expose the causal transitions that make those outcomes unfold.}
\label{tab:apt_chain_examples_app}
\small
\setlength{\tabcolsep}{4pt}
\begin{tabular}{@{}p{3.0cm}p{6.6cm}p{5.8cm}@{}}
\toprule
\textbf{Coarse event label} & \textbf{Possible APT chain} & \textbf{Causal interpretation} \\
\midrule
Ball bounces on a table &
\texttt{contact\_init} $\rightarrow$ \texttt{elastic\_rebound} $\rightarrow$ \texttt{contact\_break} $\rightarrow$ \texttt{landing} $\rightarrow$ \texttt{settling} &
Contact activates collision response; rebound reverses motion; later support and damping stabilize the object. \\
Block falls off a platform &
\texttt{free\_fall\_onset} $\rightarrow$ \texttt{landing} $\rightarrow$ \texttt{sliding\_arrest} &
Support loss makes gravity dominate; surface contact restores support; friction dissipates sliding. \\
Puck hits a soft obstacle &
\texttt{contact\_init} $\rightarrow$ \texttt{inelastic\_capture} $\rightarrow$ \texttt{settling} &
Contact begins, impact energy is dissipated rather than rebounded, and the system stabilizes. \\
Domino falls &
\texttt{toppling} $\rightarrow$ \texttt{rotation\_onset} $\rightarrow$ \texttt{settling} &
Loss of stability induces angular motion; damping and contact eventually bring the object to equilibrium. \\
Box starts sliding and stops &
\texttt{static\_breakaway} $\rightarrow$ \texttt{sliding\_arrest} &
Applied force overcomes static friction; kinetic friction dissipates motion into rest. \\
Object crosses surfaces &
\texttt{surface\_transition} $\rightarrow$ \texttt{sliding\_arrest} &
The support condition changes, altering motion; friction on the new surface eventually stops the object. \\
\bottomrule
\end{tabular}
\end{table*}

\subsection{APT Labels as Simulation Targets}
\label{app:apt_simulation_targets}

The APT definition serves two roles.
First, it tells annotators what transition to mark in an existing video.
Second, it provides a label-first specification for simulation.
Once an APT label or chain is fixed, a simulator can be configured to instantiate that transition by controlling geometry, mass, initial velocity, applied force, friction, restitution, support, and layout.
For example, an \texttt{elastic\_rebound} scene can be produced by setting a rigid barrier and sufficient restitution; a \texttt{sliding\_arrest} scene can be produced by setting an initial sliding velocity and frictional surface; a \texttt{toppling} scene can be produced by applying an off-center push to a tall object near its stability boundary.

This label-first route is important because APTs need not be discovered passively.
Rare transitions can be generated by specifying the desired causal pattern before rendering the scene.
Thus the same label space functions as an annotation unit, an evaluation target, a simulation specification, and a training supervision signal.

\subsection{Mechanism-Conditioned Decoding}
\label{app:apt_mechanism_decoding}

APT prediction can be formulated as mechanism-conditioned decoding.
Instead of mapping video evidence directly to one of all transition labels, the model first identifies the active physical domain and then decodes the transition type within that domain:
\begin{equation}
    p(\ell_i \mid V, \tau_i)
    =
    \sum_{d_i}
    p(\ell_i \mid d_i, V, \tau_i)\,
    p(d_i \mid V, \tau_i).
    \label{eq:mechanism_conditioned_app}
\end{equation}
This factorization reflects the causal organization of the taxonomy.
Support loss and downward motion point to the gravity domain before \texttt{free\_fall\_onset};
gap closure points to the contact domain before \texttt{contact\_init};
approach-to-retreat reversal after impact points to collision response before \texttt{elastic\_rebound};
sliding-to-rest motion points to friction before \texttt{sliding\_arrest};
loss of balance points to stability before \texttt{toppling}.

This structure is useful for learning because it discourages pure label memorization.
The model must associate a visible cue with a physical mechanism and then map that mechanism to a transition type.
In the main method, this idea is implemented through mechanism-conditioned APT prompts and format-conditional co-training.
The causal scaffold encourages the model to learn APTs as reusable physical representations rather than as a narrow JSON output schema.

\subsection{Summary of the Supplementary Definition}
\label{app:apt_summary}

APTs are defined to capture the causal transitions that event-level labels collapse.
They are visually grounded, vocabulary-level atomic, and tied to physical mechanisms.
A video is represented as an ordered APT chain, but the chain can be expressed in multiple formats, including JSON, QA, multiple-choice, and natural-language description.
The taxonomy is deliberately scoped to macroscopic rigid-body mechanics, where transitions are both physically meaningful and visually localizable.
This makes APTs suitable not only for diagnosis, but also for simulation construction and format-robust causal fine-tuning.

\section{Supplementary Details: Rule-Guided Coarse-to-Fine APT Construction}
\label{supp:rule_guided_apt_construction}

This section provides implementation details for our rule-guided coarse-to-fine APT construction pipeline.
We use human annotations as the perceptual validation anchor, \texttt{qwen/Qwq-32B} as the text-only LLM for category-level scene proposal and GT-trace parsing, and \texttt{Qwen3.5-VL} as the VLM for post-render semantic grounding.
The central design principle is to use each component for the part of the pipeline where it is most reliable: humans provide physical common sense, LLMs expand and parse structured templates, coding agents make templates executable, simulators execute physics and record ground-truth state traces, and VLMs identify concrete visual objects after rendering.
Thus, LLMs and VLMs are not treated as the source of physical truth.
Final APT labels are determined from realized simulator traces and calibrated against human-visible transitions.

\subsection{Human-Labeled Validation Anchor}
\label{supp:human_anchor}

We use human annotation as the perceptual anchor for APT construction.
Although simulation provides scalable state traces, APTs are visually grounded physical transitions, so their boundaries must match what humans can reliably perceive in rendered video.
We annotate controlled physical videos from CLEVRER~\citep{yi2020clevrer} and Physion++~\citep{bear2021physion,tung2023physionpp}: annotators watch each clip, mark visible physical state changes, assign one of the 14 APT types, and record transition timestamps.
The human-labeled anchor contains 500 CLEVRER validation trials with 6{,}316 APTs, covering 7 of 14 types, and 746 Physion++ trials with 20{,}987 APTs from six rigid-body scenario families, yielding 1{,}246 trials, 27{,}303 APTs, and full 14-type coverage in total.
We use this anchor to calibrate coupling rules, prompts, YAML success checks, simulator-trace parsing rules, and motion/contact thresholds.

\subsection{Rule-Guided Coarse-to-Fine Pipeline}
\label{supp:pipeline}

APT types are atomic, but they are not freely composable.
Randomly combining transition labels often produces physically incoherent or visually unnatural scenes.
For example, gravity-driven falling naturally couples with landing, contact initiation, rebound, contact break, and settling; friction naturally couples with sliding and arrest; and stability loss naturally couples with rotation, toppling, landing, and settling.
We therefore use a rule-guided construction process rather than random APT sampling.
Humans first define admissible causal couplings; an LLM expands them into category-level scene templates under simulator constraints; a coding agent converts templates into executable YAML; asynchronous domain-randomized simulation produces rendered videos and state traces; a VLM grounds the rendered appearance; and an LLM-assisted parser converts simulator traces into realized APT chains.

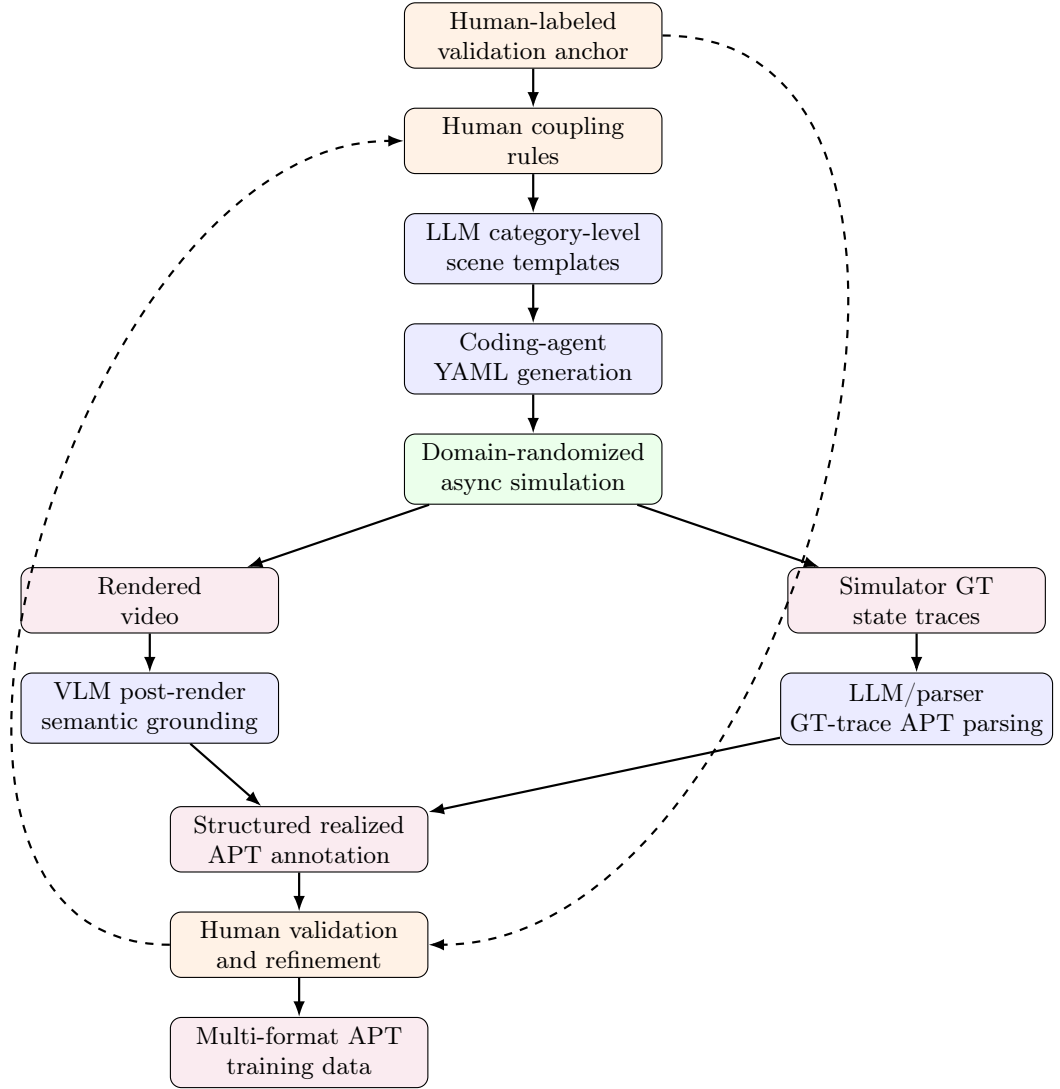
\begin{figure}[htbp]
\centering
\resizebox{0.98\linewidth}{!}{
\begin{tikzpicture}[
    node distance=5mm and 11mm,
    every node/.style={font=\small},
    box/.style={draw, rounded corners, align=center, minimum width=33mm, minimum height=8mm, fill=gray!5},
    human/.style={box, fill=orange!10},
    model/.style={box, fill=blue!8},
    sim/.style={box, fill=green!8},
    data/.style={box, fill=purple!8},
    arrow/.style={-{Latex[length=2mm]}, thick}
]
\node[human] (anchor) {Human-labeled\\validation anchor};
\node[human, below=of anchor] (rules) {Human coupling\\rules};
\node[model, below=of rules] (llmtemp) {LLM category-level\\scene templates};
\node[model, below=of llmtemp] (yaml) {Coding-agent\\YAML generation};
\node[sim, below=of yaml] (sim) {Domain-randomized\\async simulation};

\node[data, below left=8mm and 16mm of sim] (video) {Rendered\\video};
\node[data, below right=8mm and 16mm of sim] (trace) {Simulator GT\\state traces};

\node[model, below=of video] (vlm) {VLM post-render\\semantic grounding};
\node[model, below=of trace] (parser) {LLM/parser\\GT-trace APT parsing};

\node[data, below right=8mm and -14mm of vlm] (apt) {Structured realized\\APT annotation};
\node[human, below=of apt] (calib) {Human validation\\and refinement};
\node[data, below=of calib] (multi) {Multi-format APT\\training data};

\draw[arrow] (anchor) -- (rules);
\draw[arrow] (rules) -- (llmtemp);
\draw[arrow] (llmtemp) -- (yaml);
\draw[arrow] (yaml) -- (sim);
\draw[arrow] (sim) -- (video);
\draw[arrow] (sim) -- (trace);
\draw[arrow] (video) -- (vlm);
\draw[arrow] (trace) -- (parser);
\draw[arrow] (vlm) -- (apt);
\draw[arrow] (parser) -- (apt);
\draw[arrow] (apt) -- (calib);
\draw[arrow] (calib) -- (multi);
\draw[arrow, dashed] (calib.west) to[out=180,in=180] (rules.west);
\draw[arrow, dashed] (anchor.east) to[out=0,in=0] (calib.east);
\end{tikzpicture}
}
\caption{\textbf{Rule-guided coarse-to-fine APT construction.}
Humans provide a perceptual anchor and admissible causal couplings.
An LLM proposes category-level templates, a coding agent converts them to executable YAML, and asynchronous simulation executes them under domain randomization.
After rendering, a VLM grounds the concrete visible objects, while an LLM-assisted parser reads simulator GT traces to recover precise APT stages.
Human validation refines prompts, templates, thresholds, and success checks, yielding multi-format supervision.}
\label{fig:supp_rule_guided_pipeline}
\end{figure}

\subsection{Step 1: Human Coupling Rules}
\label{supp:human_coupling}

Humans first specify coarse physical couplings rather than complete videos or frame-level labels.
Each rule describes which APT domains can naturally co-occur and in what approximate temporal order.
This prevents the generator from sampling arbitrary APT sequences that have no plausible physical realization.

\begin{lstlisting}[style=answerstyle, caption={Demo human-defined coupling rule.}]
Allowed domains:
  friction, contact, gravity, settling

Allowed APT types:
  static_breakaway
  surface_transition
  sliding_arrest
  free_fall_onset
  contact_init
  landing
  inelastic_capture
  settling

Physical sense:
  A dynamic rigid object can slide on a support surface, cross into a rougher region, and stop due to friction.
  A garment can become unsupported, fall under gravity, contact the rigid object or support surface, and settle without rebound.
  These transitions are physically compatible and can form a longer APT chain than a simple single-event clip.

Simulator asset classes allowed:
  dynamic_rigid, static_rigid, articulated, garment, fluid, sand
\end{lstlisting}

\subsection{Step 2: LLM Category-Level Scene Template Proposal}
\label{supp:llm_template}

Given a human coupling rule, \texttt{qwen/Qwq-32B} proposes category-level scene templates.
The LLM is not asked to choose final assets or annotate timestamps.
Instead, it proposes a causal scene skeleton under simulator constraints: which object classes participate, which interactions should occur, and which APT domains are expected.

\begin{lstlisting}[style=promptstyle, caption={Prompt for LLM category-level scene-template proposal.}]
You are helping construct simulation scenes for Atomic Physical Transitions (APTs).
The goal is to create physically plausible scenes that contain multiple coupled APTs.

Supported simulator asset classes:
  dynamic_rigid
  static_rigid
  articulated
  garment
  fluid
  sand

Human-defined coupling rule:
  Friction, contact, gravity, and settling may co-occur.
  A dynamic rigid object may slide on a support surface and stop due to friction.
  A garment may fall onto the rigid object or support surface and then settle.

Target APT domains:
  friction, contact, gravity, settling

Instructions:
  Propose 3 category-level scene templates.
  For each template, output:
    (1) scene name,
    (2) object slots and their simulator asset classes,
    (3) natural-language causal scene skeleton,
    (4) expected APT types,
    (5) success checks,
    (6) likely failure cases.

Do not choose final object identity, color, or material.
Do not output timestamps.
\end{lstlisting}

\begin{lstlisting}[style=answerstyle, caption={Demo LLM answer: category-level templates.}]
Template 1: rigid-slide-garment-cover

Object slots:
  support: static_rigid, e.g., table or platform
  mover: dynamic_rigid, e.g., cube, sphere, cylinder, mug, or apple
  cover: garment, e.g., pants, shorts, or cloth

Causal scene skeleton:
  A dynamic rigid object starts on a support surface, receives an initial horizontal velocity,
  slides across the surface, crosses into a higher-friction region, and stops.
  A garment is released above the rigid object, falls under gravity, contacts the rigid object
  or the support surface, and settles.

Expected APTs:
  static_breakaway
  surface_transition
  sliding_arrest
  free_fall_onset
  contact_init
  landing
  inelastic_capture
  settling

Success checks:
  The rigid object should move tangentially for at least 0.3 s and later remain below a velocity threshold.
  The garment should have a clear downward motion phase, create a new contact pair, and remain approximately at rest.

Failure cases:
  The rigid object never moves.
  The garment misses the support area.
  The garment penetrates the table.
  The garment remains airborne at the end.
  The rigid object rebounds instead of stopping.

Template 2: rigid-bump-then-cloth-damping

Object slots:
  support: static_rigid
  mover: dynamic_rigid
  obstacle: static_rigid or dynamic_rigid
  cover: garment

Causal scene skeleton:
  A dynamic rigid object slides into a low obstacle, changes direction or starts rotating,
  slows on a rough surface, and is later covered by a falling garment that damps residual motion.

Expected APTs:
  static_breakaway
  contact_init
  rotation_onset
  sliding_arrest
  free_fall_onset
  landing
  settling

Template 3: falling-rigid-with-garment-capture

Object slots:
  support: static_rigid
  falling_object: dynamic_rigid
  cover: garment

Causal scene skeleton:
  A small dynamic rigid object falls from above a support surface, lands with low restitution,
  slides briefly, and is later covered by a falling garment.

Expected APTs:
  free_fall_onset
  landing
  contact_init
  inelastic_capture
  sliding_arrest
  settling
\end{lstlisting}

\subsection{Step 3: Coding-Agent YAML Generation}
\label{supp:yaml_generation}

A coding agent converts the selected category-level scene template into an executable YAML scene.
This step makes the template simulatable and introduces diversity in initial conditions.
The YAML specifies object slots, candidate assets, randomized positions, initial velocities, material ranges, friction and restitution ranges, camera and lighting settings, success checks, and trace logging.
All asset paths below are anonymized.

\begin{lstlisting}[style=promptstyle, caption={Prompt for coding-agent YAML generation.}]
Convert the following category-level APT template into an executable YAML scene.

Template name:
  rigid-slide-garment-cover

Scene skeleton:
  A dynamic rigid object slides on a table, crosses into a rougher region,
  slows due to friction, and stops.
  A garment is released above the rigid object, falls onto it or the table,
  and settles.

Available anonymized asset roots:
  Tables:
    ~/Desk/...
  Dynamic rigid objects:
    ~/Object/Mug
    ~/Object/apple
    ~/Object/Sugar_Box
  Garments:
    ~/Object/Garment/Pants
  Materials:
    ~/Material/...
  Backgrounds:
    ~/Background/...

Requirements:
  Use domain randomization over asset identity, color, material, initial pose,
  initial velocity, friction, restitution, lighting, and camera.
  Record per-frame pose, linear velocity, angular velocity, contact pairs,
  support relation, object identity, material parameters, and surface region ID.
  Add success checks for sliding motion, surface transition, arrest, garment falling,
  garment contact, and final settling.
  Return YAML only.
\end{lstlisting}

\begin{lstlisting}[style=yamlstyle, caption={Demo coding-agent YAML: friction-contact-garment scene.}]
scene_name: rigid_slide_garment_cover
description: >
  A dynamic rigid object slides on a table, enters a higher-friction region,
  stops, and is later covered by a falling garment.

background:
  intensity: [1000, 2000]
  texture:
    - "~/Background/champagne_castle_1_4k.hdr"
    - "~/Background/evening_road_01_4k.hdr"
    - "~/Background/kloofendal_48d_partly_cloudy_4k.hdr"
    - "~/Background/kloppenheim_02_4k.hdr"
    - "~/Background/mealie_road_4k.hdr"
    - "~/Background/moonlit_golf_4k.hdr"
    - "~/Background/noon_grass_4k.hdr"
    - "~/Background/qwantani_4k.hdr"

light:
  light_1:
    types: ["Rect", "Sphere", "Cylinder"]
    common:
      initial_pos_range: [-0.2, -0.2, 0.0, 0.2, 0.5, 0.0]
      initial_ori_range: [-30, -30, -30, 30, 30, 30]
      color_temperature_range: [3000, 9000]
      color_range: [[0.8, 0.8, 0.8], [1.2, 1.2, 1.2]]
    Rect:
      intensity_range: [2000, 5000]
      width_range: [1.0, 3.0]
      height_range: [1.0, 3.0]
    Sphere:
      intensity_range: [5000, 10000]
      radius_range: [0.1, 0.5]
    Cylinder:
      intensity_range: [4000, 8000]
      radius_range: [0.1, 0.3]
      length_range: [1.0, 2.0]

camera:
  random: true
  look_at: [0.0, 0.2, 0.75]
  pos_range: [-1.8, -2.2, 1.2, 1.8, -1.2, 2.2]
  focal_length_range: [24, 45]

objects:
  support_table:
    usd:
      - "~/Desk/Collected_appleseed_coffeetable/appleseed_coffeetable_inst_base.usd"
      - "~/Desk/Collected_cline/cline_inst_base.usd"
      - "~/Desk/Collected_oaktablesmall/oaktablesmall_inst_base.usd"
      - "~/Desk/Collected_willowbench/willowbench_inst_base.usd"
    random: true
    default_material: true
    num_per_env: 1
    semantic_label: "static_support_surface"
    support_table_1:
      pos: [0.0, 0.0, 0.0]
      ori: [0.0, 0.0, 90.0]
      visual:
        scale: [2.0, 2.0, 1.0]
        color:
          - [0.8, 0.7, 0.6]
          - [0.6, 0.5, 0.4]
          - [0.4, 0.3, 0.2]
          - [0.2, 0.1, 0.0]
        visual_material:
          mdl_folder: "~/Material/Base/Textiles"
      physics:
        type: "geometry"
        collision: true
        surface_regions:
          low_friction_region:
            x_range: [-0.70, -0.05]
            static_friction_range: [0.05, 0.20]
            dynamic_friction_range: [0.05, 0.20]
          high_friction_region:
            x_range: [-0.05, 0.70]
            static_friction_range: [0.80, 1.30]
            dynamic_friction_range: [0.70, 1.20]

  mover:
    common:
      primitive_scale: [0.30, 0.30, 0.30]
      initial_pos_range: [-0.60, -0.05, 0.75, -0.45, 0.20, 0.82]
      initial_ori_range: [-10, -10, -20, 10, 10, 20]
    usd:
      - "~/Object/Mug"
      - "~/Object/apple"
      - "~/Object/Sugar_Box"
    random: true
    num_per_env: 1
    semantic_label: "dynamic_rigid_mover"
    mover_1:
      visual:
        scale: [1.0, 1.0, 1.0]
        visible: true
        color:
          - [1.0, 0.0, 0.0]
          - [0.0, 0.0, 1.0]
          - [1.0, 1.0, 0.0]
          - [0.0, 1.0, 1.0]
        visual_material:
          mdl_folder: "~/Material/Base/Textiles"
      physics:
        type: "dynamic"
        collision: true
        mass_range: [0.2, 0.8]
        ratio: 1.1
        physics_material:
          static_friction_range: [0.10, 0.35]
          dynamic_friction_range: [0.10, 0.35]
          restitution_range: [0.0, 0.10]
        linear_velocity_range:
          - [0.9, -0.05, 0.0]
          - [1.4, 0.05, 0.0]
        angular_velocity_range:
          - [0.0, 0.0, -0.5]
          - [0.0, 0.0, 0.5]
        track_contact_forces: true
        prepare_contact_sensor: true

  garment_cover:
    common:
      primitive_scale: [0.50, 0.50, 0.50]
      initial_pos_range: [0.05, -0.10, 1.35, 0.35, 0.25, 1.70]
      initial_ori_range: [-15, -15, 60, 15, 15, 120]
    usd:
      - "~/Object/Garment/Pants"
    random: true
    num_per_env: 1
    semantic_label: "falling_garment"
    garment_cover_1:
      visual:
        scale: [1.0, 1.0, 1.0]
        visible: true
        color:
          - [1.0, 0.0, 0.0]
          - [0.0, 1.0, 0.0]
          - [0.0, 0.0, 1.0]
          - [1.0, 1.0, 0.0]
          - [1.0, 0.0, 1.0]
          - [0.0, 1.0, 1.0]
        visual_material:
          mdl_folder: "~/Material/Base/Textiles"
      physics:
        type: "garment"
        collision: true
        ratio: 1.1
        release_time_range: [0.55, 0.90]
        particle_system:
          particle_system_enabled: true
          enable_ccd: true
          solver_position_iteration_count: 16
          max_depenetration_velocity: null
          global_self_collision_enabled: true
          non_particle_collision_enabled: true
          contact_offset: 0.010
          rest_offset: 0.006
          particle_contact_offset: 0.010
          fluid_rest_offset: 0.006
          solid_rest_offset: 0.006
        particle_material:
          adhesion: 10
          adhesion_offset_scale: 0.0
          cohesion: 0.1
          particle_adhesion_scale: 3.0
          particle_friction_scale: 3.0
          drag: 0.0
          lift: 0.0
          friction: 10.0
          damping: 5.0
          gravity_scale: 1.0
        garment_config:
          particle_mass: 5e-10
          self_collision: true
          self_collision_filter: true
          stretch_stiffness: 1e12
          bend_stiffness: 1e3
          shear_stiffness: 1e3
          spring_damping: 5.0

trace_logging:
  fps: 60
  fields:
    - pose
    - linear_velocity
    - angular_velocity
    - contact_pairs
    - contact_forces
    - support_relation
    - object_identity
    - material_parameters
    - surface_region_id

success_checks:
  intended_domains:
    - friction
    - contact
    - gravity
    - settling
  required_conditions:
    mover_slides:
      object: "mover"
      min_horizontal_displacement: 0.25
      min_duration_sec: 0.25
    surface_transition:
      object: "mover"
      from_region: "low_friction_region"
      to_region: "high_friction_region"
    sliding_arrest:
      object: "mover"
      speed_below: 0.03
      hold_duration_sec: 0.20
    garment_falls:
      object: "garment_cover"
      min_downward_displacement: 0.30
    garment_contacts:
      object: "garment_cover"
      target_any_of: ["mover", "support_table"]
    final_settling:
      objects: ["mover", "garment_cover"]
      linear_speed_below: 0.03
      angular_speed_below: 0.10
      hold_duration_sec: 0.30
  failure_filters:
    - "garment misses table and falls out of view"
    - "mover never enters high-friction region"
    - "mover leaves support surface"
    - "simulation contains severe interpenetration"
\end{lstlisting}

\subsection{Step 4: Domain-Randomized Asynchronous Simulation}
\label{supp:dr_async_sim}

We execute each YAML scene in Isaac Lab~\citep{mittal2025isaaclab} with asynchronous parallel simulation and domain randomization.
The same category-level template can produce many concrete videos because asset identity, color, material, physical parameters, initial pose, initial velocity, camera, and lighting are sampled at simulation time.
For example, the abstract slot \texttt{dynamic\_rigid\_mover} may render as a blue cube, a red mug, a yellow cylinder, or a green apple; the \texttt{falling\_garment} slot may render as red pants, blue pants, or a cloth-like garment.
This design increases both visual diversity and APT-coupling diversity while preserving the intended causal structure.

The simulator records two synchronized outputs:
(1) rendered video frames and
(2) frame-level ground-truth state traces, including object identity, pose, linear velocity, angular velocity, contact pairs, contact forces, support relations, material parameters, and surface-region identifiers.
The intended coupling rule is only a causal prior.
The final labels are assigned from the realized trace, so failed executions can be filtered, unintended outcomes can be relabeled, and extra realized transitions can be retained when supported by the trace.

\subsection{Step 5: VLM Post-Render Semantic Grounding}
\label{supp:vlm_grounding}

Because domain randomization chooses concrete assets during simulation, the abstract template does not know the final visible objects.
We therefore use \texttt{Qwen3.5-VL} after rendering to ground object slots into concrete visual descriptions.
The VLM identifies object category, color, material appearance, location, and visible relations.
It is not asked to infer APT types or timestamps.

\begin{lstlisting}[style=promptstyle, caption={Prompt for VLM post-render semantic grounding.}]
You are given frames from a rendered physics simulation and the abstract scene template below.

Abstract slots:
  support: a static support surface
  mover: a dynamic rigid object sliding on the support
  cover: a garment released above the mover

Task:
  Identify the concrete visible object corresponding to each slot.
  Return only JSON with:
    slot
    object_category
    color
    material_appearance
    location
    visible_relation

Important:
  Do not infer APT labels.
  Do not infer timestamps.
  Do not describe invisible simulator states.
  Only ground the abstract template into visible objects and relations.
\end{lstlisting}

\begin{lstlisting}[style=answerstyle, caption={Demo VLM answer: semantic grounding.}]
{
  "support": {
    "object_category": "coffee table",
    "color": "brown",
    "material_appearance": "wood-like",
    "location": "center of the scene",
    "visible_relation": "supports the moving object and later the garment"
  },
  "mover": {
    "object_category": "cube",
    "color": "blue",
    "material_appearance": "matte rigid object",
    "location": "on top of the table",
    "visible_relation": "slides from left to right across the table"
  },
  "cover": {
    "object_category": "pair of pants",
    "color": "red",
    "material_appearance": "soft textile garment",
    "location": "above the table, then on or near the cube",
    "visible_relation": "falls downward and contacts or covers the cube"
  }
}
\end{lstlisting}

\subsection{Step 6: LLM/Parser GT-Trace-Based APT Stage Parsing}
\label{supp:gt_trace_parsing}

Precise APT timing is obtained from simulator traces, not from visual guessing.
We use \texttt{qwen/Qwq-32B} as an LLM-assisted structured parser.
The parser receives the APT taxonomy, the human-defined coupling rule, the VLM-grounded object names, and a compact trace summary.
It converts numeric state changes into realized APT instances.
The intended template is treated as a prior, not a label guarantee.

\begin{lstlisting}[style=promptstyle, caption={Prompt for LLM GT-trace-to-APT parsing.}]
You are converting simulator ground-truth traces into Atomic Physical Transition annotations.

Important rules:
  Use simulator traces as the source of timing and physical state.
  Use VLM grounding only for object names and visual descriptions.
  The intended template is a prior, not a label guarantee.
  If the realized dynamics differ from the intended template, relabel according to the trace.
  Output concise evidence, not chain-of-thought.

APT definitions:
  contact_init: free then contact; a new object-object or object-surface contact begins.
  contact_break: contact then free; an existing contact pair separates.
  elastic_rebound: approach then retreat; post-contact relative velocity reverses with rebound.
  inelastic_capture: moving then captured/resting; object remains in contact without rebound.
  free_fall_onset: supported then falling; object becomes unsupported and moves downward.
  projectile_apex: ascending then descending; vertical velocity crosses from positive to negative.
  landing: falling then supported; falling object first becomes supported by a surface/object.
  static_breakaway: stationary then sliding; object begins sliding after force overcomes static friction.
  sliding_arrest: sliding then stationary; sliding object comes to rest and remains below threshold.
  surface_transition: moving object crosses from one surface/material region to another.
  rotation_onset: non-rotating then rotating; angular velocity becomes visible/significant.
  toppling: stable then unstable; object loses stability and begins falling over.
  rotation_arrest: rotating then stationary; angular velocity damps below threshold.
  settling: object motion damps to stable rest after wobble, bounce, sliding, or deformation.

Grounded slots:
  mover = blue cube
  support = brown wooden table
  cover = red pants

Trace summary:
  fps = 60

  t=0.00s:
    blue_cube supported_by table = true
    blue_cube speed = 0.00 m/s

  t=0.15s:
    blue_cube horizontal impulse applied
    blue_cube speed increases from 0.00 to 1.08 m/s

  t=0.44s:
    blue_cube x crosses surface_region low_friction -> high_friction
    speed begins decreasing more rapidly

  t=0.81s:
    blue_cube speed < 0.03 m/s for 0.20s
    blue_cube remains supported_by table = true

  t=0.88s:
    red_pants release flag changes fixed -> dynamic
    red_pants supported = false
    red_pants vertical velocity becomes negative

  t=1.18s:
    contact_pair(red_pants, blue_cube) changes false -> true
    contact_force normal > threshold
    red_pants vertical velocity magnitude decreases sharply

  t=1.25s:
    red_pants remains in contact with blue_cube/table
    no rebound; relative normal velocity does not become positive

  t=1.70s:
    red_pants particle velocities < threshold for 0.30s
    blue_cube speed remains < threshold

Output format:
  A JSON array.
  Each record must contain:
    type
    t_ms
    objects
    evidence
\end{lstlisting}

\begin{lstlisting}[style=answerstyle, caption={Demo LLM answer: realized APT chain.}]
[
  {
    "type": "static_breakaway",
    "t_ms": 150,
    "objects": ["blue cube"],
    "evidence": "blue cube changes from rest to horizontal sliding after impulse"
  },
  {
    "type": "surface_transition",
    "t_ms": 440,
    "objects": ["blue cube", "brown wooden table"],
    "evidence": "blue cube crosses from low-friction to high-friction surface region"
  },
  {
    "type": "sliding_arrest",
    "t_ms": 810,
    "objects": ["blue cube", "brown wooden table"],
    "evidence": "blue cube speed stays below threshold while supported by table"
  },
  {
    "type": "free_fall_onset",
    "t_ms": 880,
    "objects": ["red pants"],
    "evidence": "red pants become unsupported and acquire downward velocity"
  },
  {
    "type": "contact_init",
    "t_ms": 1180,
    "objects": ["red pants", "blue cube"],
    "evidence": "new contact pair between red pants and blue cube appears"
  },
  {
    "type": "landing",
    "t_ms": 1180,
    "objects": ["red pants", "blue cube"],
    "evidence": "falling red pants first become supported through contact"
  },
  {
    "type": "inelastic_capture",
    "t_ms": 1250,
    "objects": ["red pants", "blue cube"],
    "evidence": "red pants remain in contact without rebound after impact"
  },
  {
    "type": "settling",
    "t_ms": 1700,
    "objects": ["red pants", "blue cube"],
    "evidence": "garment and cube remain below motion thresholds for the hold window"
  }
]
\end{lstlisting}

This example illustrates why rule-guided simulation produces more compositionally dense APT data than existing event-centered clips.
A conventional falling or bouncing clip may contain only two or three obvious transitions.
In contrast, the coupled friction-contact-garment scene produces a longer chain involving sliding, surface transition, arrest, falling, contact, landing, capture, and settling.

\subsection{Additional Demo: Gravity-Contact Rigid-Body Scene}
\label{supp:gravity_contact_demo}

We also construct simpler rigid-body templates to cover gravity, contact, landing, inelastic capture, and settling.
The example below follows the same rule-guided pipeline but uses only static rigid supports and dynamic rigid objects.

\begin{lstlisting}[style=promptstyle, caption={Prompt for LLM scene-template proposal: gravity-contact rigid bodies.}]
Human coupling rule:
  Gravity, contact, landing, inelastic capture, and settling can co-occur.
  Use only static rigid supports and dynamic rigid objects.
  Construct a category-level scene where multiple small rigid objects fall onto a table
  and may contact one another after landing.

Output:
  object slots
  scene skeleton
  expected APTs
  success checks
  likely failure cases
\end{lstlisting}

\begin{lstlisting}[style=answerstyle, caption={Demo LLM answer: gravity-contact template.}]
{
  "template_name": "multi_rigid_drop_on_table",
  "object_slots": {
    "support": "static_rigid table",
    "falling_object_A": "dynamic_rigid object",
    "falling_object_B": "dynamic_rigid object"
  },
  "scene_skeleton": [
    "Two dynamic rigid objects are initialized above a table.",
    "They fall under gravity.",
    "Each object lands on the table.",
    "One object may roll or slide into the other.",
    "Both objects settle on the table."
  ],
  "expected_apts": [
    "free_fall_onset",
    "landing",
    "contact_init",
    "inelastic_capture",
    "sliding_arrest",
    "settling"
  ],
  "success_checks": [
    "each dynamic object has downward velocity before first table contact",
    "each object creates a new contact pair with the table",
    "at least one post-landing object-object contact occurs or the scene is relabeled",
    "all dynamic objects satisfy final rest thresholds"
  ],
  "likely_failure_cases": [
    "one object misses the table",
    "objects bounce too strongly and leave the view",
    "objects never contact each other",
    "final settling is not reached within the clip"
  ]
}
\end{lstlisting}

\begin{lstlisting}[style=yamlstyle, caption={Demo coding-agent YAML: gravity-contact rigid-body scene.}]
scene_name: multi_rigid_drop_on_table
description: >
  Two dynamic rigid objects fall onto a table, land, may collide with
  each other, and finally settle.

background:
  intensity: [1000, 2000]
  texture:
    - "~/Background/champagne_castle_1_4k.hdr"
    - "~/Background/evening_road_01_4k.hdr"
    - "~/Background/noon_grass_4k.hdr"
    - "~/Background/qwantani_4k.hdr"

light:
  light_1:
    types: ["Rect", "Sphere", "Cylinder"]
    common:
      initial_pos_range: [-0.2, -0.2, 0.0, 0.2, 0.5, 0.0]
      initial_ori_range: [-30, -30, -30, 30, 30, 30]
      color_temperature_range: [3000, 9000]
      color_range: [[0.8, 0.8, 0.8], [1.2, 1.2, 1.2]]
    Rect:
      intensity_range: [2000, 5000]
      width_range: [1.0, 3.0]
      height_range: [1.0, 3.0]
    Sphere:
      intensity_range: [5000, 10000]
      radius_range: [0.1, 0.5]
    Cylinder:
      intensity_range: [4000, 8000]
      radius_range: [0.1, 0.3]
      length_range: [1.0, 2.0]

camera:
  random: true
  look_at: [0.2, 0.4, 0.75]
  pos_range: [-1.5, -2.0, 1.2, 1.5, -1.0, 2.2]
  focal_length_range: [24, 45]

objects:
  simple_desk:
    usd:
      - "~/Desk/Collected_appleseed_coffeetable/appleseed_coffeetable_inst_base.usd"
      - "~/Desk/Collected_cline/cline_inst_base.usd"
      - "~/Desk/Collected_oaktablesmall/oaktablesmall_inst_base.usd"
      - "~/Desk/Collected_willowbench/willowbench_inst_base.usd"
    random: true
    default_material: true
    num_per_env: 1
    semantic_label: "static_support_surface"
    simple_desk_1:
      pos: [0.0, 0.0, 0.0]
      ori: [0.0, 0.0, 90.0]
      visual:
        scale: [1.3, 1.3, 1.0]
        color:
          - [0.8, 0.7, 0.6]
          - [0.6, 0.5, 0.4]
          - [0.4, 0.3, 0.2]
          - [0.2, 0.1, 0.0]
        visual_material:
          material_usd_folder: "~/Material/Garment"
      physics:
        type: "geometry"
        collision: true
        physics_material:
          static_friction_range: [0.5, 1.0]
          dynamic_friction_range: [0.4, 0.9]
          restitution_range: [0.0, 0.15]

  falling_rigid_A:
    common:
      primitive_scale: [0.30, 0.30, 0.30]
      initial_pos_range: [-0.35, -0.20, 1.30, -0.05, 0.20, 1.80]
      initial_ori_range: [-30, -30, -30, 30, 30, 30]
    usd:
      - "~/Object/Mug"
      - "~/Object/apple"
      - "~/Object/Sugar_Box"
    random: true
    num_per_env: 1
    semantic_label: "falling_rigid_A"
    falling_rigid_A_1:
      visual:
        scale: [1.0, 1.0, 1.0]
        visible: true
        color:
          - [1.0, 0.0, 0.0]
          - [0.0, 0.0, 1.0]
          - [1.0, 1.0, 0.0]
          - [0.0, 1.0, 1.0]
        visual_material:
          material_usd_folder: "~/Material/Garment"
      physics:
        type: "dynamic"
        collision: true
        mass_range: [0.2, 0.7]
        ratio: 1.1
        physics_material:
          static_friction_range: [0.2, 0.6]
          dynamic_friction_range: [0.2, 0.6]
          restitution_range: [0.0, 0.20]
        linear_velocity_range:
          - [0.05, 0.00, 0.00]
          - [0.25, 0.10, 0.00]
        angular_velocity_range:
          - [-0.2, -0.2, -0.2]
          - [0.2, 0.2, 0.2]
        track_contact_forces: true
        prepare_contact_sensor: true

  falling_rigid_B:
    common:
      primitive_scale: [0.30, 0.30, 0.30]
      initial_pos_range: [0.05, 0.00, 1.25, 0.35, 0.30, 1.75]
      initial_ori_range: [-30, -30, -30, 30, 30, 30]
    usd:
      - "~/Object/Mug"
      - "~/Object/apple"
      - "~/Object/Sugar_Box"
    random: true
    num_per_env: 1
    semantic_label: "falling_rigid_B"
    falling_rigid_B_1:
      visual:
        scale: [1.0, 1.0, 1.0]
        visible: true
        color:
          - [0.0, 1.0, 0.0]
          - [1.0, 0.0, 1.0]
          - [0.0, 1.0, 1.0]
          - [1.0, 0.5, 0.0]
        visual_material:
          material_usd_folder: "~/Material/Garment"
      physics:
        type: "dynamic"
        collision: true
        mass_range: [0.2, 0.7]
        ratio: 1.1
        physics_material:
          static_friction_range: [0.2, 0.6]
          dynamic_friction_range: [0.2, 0.6]
          restitution_range: [0.0, 0.20]
        linear_velocity_range:
          - [-0.25, -0.10, 0.00]
          - [-0.05, 0.00, 0.00]
        angular_velocity_range:
          - [-0.2, -0.2, -0.2]
          - [0.2, 0.2, 0.2]
        track_contact_forces: true
        prepare_contact_sensor: true

trace_logging:
  fps: 60
  fields:
    - pose
    - linear_velocity
    - angular_velocity
    - contact_pairs
    - contact_forces
    - support_relation
    - object_identity
    - material_parameters

success_checks:
  intended_domains:
    - gravity
    - contact
    - friction
    - settling
  required_conditions:
    object_A_falls:
      object: "falling_rigid_A"
      min_downward_velocity: 0.2
    object_B_falls:
      object: "falling_rigid_B"
      min_downward_velocity: 0.2
    table_landing_A:
      contact_pair: ["falling_rigid_A", "simple_desk"]
    table_landing_B:
      contact_pair: ["falling_rigid_B", "simple_desk"]
    final_settling:
      objects: ["falling_rigid_A", "falling_rigid_B"]
      linear_speed_below: 0.03
      angular_speed_below: 0.10
      hold_duration_sec: 0.30
  optional_conditions:
    secondary_contact:
      contact_pair_any:
        - ["falling_rigid_A", "falling_rigid_B"]
  failure_filters:
    - "one rigid object misses the support surface"
    - "object bounces out of camera view"
    - "final settling not reached within clip"
    - "severe interpenetration"
\end{lstlisting}

\subsection{Step 7: Human Validation and Prompt/Template Refinement}
\label{supp:automatic_validation}

The automatic construction pipeline is validated against the human-labeled anchor.
For a subset of simulated clips, humans annotate the visible APT chain directly from the rendered video.
We compare human annotations with automatically produced chains using type equality and timestamp tolerance.
Disagreements are used to refine four components:
(1) coupling rules, when a proposed scene is physically unnatural;
(2) YAML success checks, when the intended transition does not reliably occur;
(3) VLM grounding prompts, when object identity or relation is ambiguous; and
(4) GT-trace parsing thresholds, when simulator state changes occur earlier or later than human-visible transitions.
After calibration, the automatic pipeline produces APT chains that closely match human annotations while scaling beyond manual labeling.

\begin{lstlisting}[style=promptstyle, caption={Prompt for validation disagreement analysis.}]
You are given:
  (1) a human-labeled APT chain,
  (2) an automatically produced APT chain,
  (3) the simulator trace summary,
  (4) the rendered-video grounding.

Compare the human and automatic chains.
For each mismatch, classify the likely cause as one of:
  template_issue
  success_check_issue
  grounding_issue
  trace_threshold_issue
  human_ambiguity
  extra_realized_transition

Suggest one concrete fix.
Do not change labels unless the simulator trace supports the change.
\end{lstlisting}

\begin{lstlisting}[style=answerstyle, caption={Demo validation answer.}]
[
  {
    "mismatch": "automatic contact_init at 1140 ms; human contact_init at 1180 ms",
    "cause": "trace_threshold_issue",
    "analysis": "contact force is nonzero before contact is visually clear",
    "fix": "increase contact-force threshold or require 2 consecutive frames above threshold"
  },
  {
    "mismatch": "automatic omitted landing for red pants",
    "cause": "success_check_issue",
    "analysis": "parser produced contact_init but did not add landing for an object that was falling immediately before first support contact",
    "fix": "if an object has negative vertical velocity before first support contact, add landing at the first support contact"
  },
  {
    "mismatch": "automatic detected extra surface_transition",
    "cause": "extra_realized_transition",
    "analysis": "simulator trace shows mover crossed from low-friction to high-friction region and the speed decay changed visibly",
    "fix": "keep the label when the motion change is visible; otherwise store it as metadata-only"
  }
]
\end{lstlisting}

\subsection{Step 8: Multi-Format APT Training Data}
\label{supp:multi_format_export}

Each accepted clip is exported into multiple aligned formats.
This is important because single-format APT JSON training can make the model an APT-format specialist and degrade ordinary event-level video answering.
The same physical scene is therefore represented as:
(1) APT JSON,
(2) grounded object description,
(3) causal transition description,
(4) event-level QA, and
(5) multiple-choice QA.
These aligned formats allow \texttt{APT-Tune} to learn APTs as reusable causal physical representations rather than a single output style.

\begin{lstlisting}[style=answerstyle, caption={Demo multi-format export for the rigid-slide-garment-cover scene.}]
APT JSON:
[
  {"type": "static_breakaway", "t_ms": 150, "objects": ["blue cube"]},
  {"type": "surface_transition", "t_ms": 440, "objects": ["blue cube"]},
  {"type": "sliding_arrest", "t_ms": 810, "objects": ["blue cube"]},
  {"type": "free_fall_onset", "t_ms": 880, "objects": ["red pants"]},
  {"type": "contact_init", "t_ms": 1180, "objects": ["red pants", "blue cube"]},
  {"type": "landing", "t_ms": 1180, "objects": ["red pants", "blue cube"]},
  {"type": "inelastic_capture", "t_ms": 1250, "objects": ["red pants", "blue cube"]},
  {"type": "settling", "t_ms": 1700, "objects": ["red pants", "blue cube"]}
]

Grounded object description:
  A blue cube slides across a brown wooden table, slows down on a rougher region,
  and stops. A red pair of pants falls from above, lands on the cube, remains in
  contact, and settles.

Causal transition description:
  The rigid object first breaks away from rest and slides. Its motion changes when
  it enters a higher-friction surface region, and friction arrests the sliding.
  The garment then becomes unsupported, falls under gravity, initiates contact with
  the cube, is captured without rebound, and settles.

Event-level QA:
  Q: What happens in the video?
  A: A blue cube slides on a table and stops, then a red garment falls onto it and settles.

Multiple-choice QA:
  Q: Which description best matches the video?
  A. The cube bounces elastically off a wall.
  B. The cube slides to a stop, and a garment falls onto it and settles.
  C. The garment flies upward after hitting the cube.
  D. The table topples over.

  Correct answer: B.
\end{lstlisting}

\subsection{Dataset Split and Evaluation Protocol}
\label{supp:evaluation_protocol}

For the human-labeled data, the train/test split is stratified by source and scenario family using seed 7.
The training split contains 997 trials, approximately 21.8K APT instances, consisting of 400 CLEVRER trials and 597 Physion++ trials.
The test split contains 249 trials, approximately 5.5K APT instances, consisting of 100 CLEVRER trials and 149 Physion++ trials.
All 14 APT types are represented, but the distribution is naturally imbalanced: \texttt{contact\_init} and \texttt{contact\_break} occur frequently in collision-heavy scenes, while \texttt{projectile\_apex} and \texttt{free\_fall\_onset} require specific geometry.
We therefore report recall and provide per-type breakdowns.

To enable fair zero-shot and fine-tuned comparison across VLMs, we use a frozen evaluation protocol.

\begin{enumerate}[leftmargin=*,topsep=2pt,itemsep=2pt]
    \item \textbf{Input normalization.}
    Each model receives the same 16 uniformly sampled frames per video at source resolution.
    Models with separate frame-rate conventions receive matched 16-frame samples.

    \item \textbf{Explicit frame times.}
    Each frame is preceded by a text marker \texttt{[Frame i/16, t=<int>ms] <image>}, where the value of \texttt{t} is computed from the frame's native index and the clip's native frame rate.
    The prompt header additionally states the frame count, clip duration, native frame rate, and valid timestamp range.
    These markers are ordinary text tokens and are identical across training, inference, and every backbone.
    Neither Qwen3-VL nor InternVL3.5 is given a family-specific video-timestamp interface, so all models receive equivalent temporal information; the same holds for the commercial APIs, which receive the identical interleaved marker/frame sequence.
    The model therefore predicts \emph{transition} times conditioned on known frame times rather than inferring frame times from pixel content.

    \item \textbf{Structured output.}
    A single shared APT-schema prompt asks the model to enumerate timed APTs as a JSON array of records:
    \texttt{\{"type": <label>, "t\_ms": <int>\}}.
    The 14 type names are listed verbatim with one-line definitions.
    No chain-of-thought is requested.

    \item \textbf{Robust parsing.}
    Predictions are parsed with a tolerant JSON parser that recovers from common formatting errors, including trailing commas, unterminated arrays, and missing closing braces.

    \item \textbf{Matching.}
    A predicted APT matches a ground-truth APT if the type matches and the timestamp lies within $\Delta t=\pm 200$\,ms.
    Matching is one-to-one and greedy by confidence when confidence is available, or by parsing order otherwise.
    Each ground-truth APT can be matched at most once.

    \item \textbf{Metrics.}
    We report recall, precision, and F1 under this one-to-one matching, together with the mean number of predicted transitions per clip, plus type accuracy on matched events and the median timing error $\Delta T$.
    At inference time predictions occasionally truncate mid-JSON when the max-token limit is reached; the tolerant parser recovers any complete leading APT records and discards the remainder rather than dropping the whole prediction, which affects fewer than 3\% of predictions.

    \item \textbf{Controlled comparison.}
    The same prompt and frame list are used for zero-shot baselines and fine-tuned models.
    The only difference across rows in the result table is the model weights.
\end{enumerate}

\subsubsection{Frame-Grid Ceiling and Tolerance Sensitivity}
\label{supp:frame_grid_ceiling}

Because the evaluation tolerance ($\pm 200$\,ms) is smaller than the spacing between uniformly sampled frames (320\,ms on CLEVRER, 806\,ms on Physion++), it is reasonable to ask whether the low zero-shot recall is an artifact of timestamp quantization rather than a failure of physical understanding.
We quantify this with the \emph{frame-grid ceiling}: the fraction of ground-truth APTs that could be matched at all if predictions were restricted to exactly the sampled-frame timestamps.
This is an upper bound on any model that only ever names a provided frame time.

\begin{table}[h]
\centering
\small
\caption{\textbf{Frame-grid ceiling (\%).} Maximum attainable recall when predicted timestamps are restricted to the sampled-frame times, as a function of frame budget and matching tolerance. At 16 frames and $\pm 200$\,ms the ceiling is 61.8\%, far above the 10--15\% observed zero-shot recall.}
\label{tab:frame_grid_ceiling}
\begin{tabular}{@{}lcccc@{}}
\toprule
\textbf{Frames} & \textbf{$\pm 100$\,ms} & \textbf{$\pm 200$\,ms} & \textbf{$\pm 400$\,ms} & \textbf{$\pm 800$\,ms} \\
\midrule
8  & 17.7 & 33.8 & 57.4 & 90.7  \\
16 & 36.0 & 61.8 & 89.0 & 100.0 \\
\bottomrule
\end{tabular}
\end{table}

Two observations follow.
First, at the reported setting (16 frames, $\pm 200$\,ms) the ceiling is 61.8\%, so quantization cannot explain zero-shot recall of 10--15\%; the binding constraint is that models fail to propose the transitions at all.
Second, under a reduced 8-frame budget, \texttt{APT-Tune} models reach 44.6--48.5\% recall, which \emph{exceeds} the corresponding 33.8\% ceiling.
A model restricted to copying provided frame times could not do this, so the tuned models interpolate transition times between the frames they are shown.
We also sweep the tolerance from $\pm 100$ to $\pm 800$\,ms and find that recall increases monotonically for every model while the ordering between models is preserved throughout, so no conclusion in the paper depends on the specific choice of $\pm 200$\,ms.
The residual limitation is genuine: a transition strictly briefer than the frame spacing may leave no visible evidence in any sampled frame, and such cases are counted as misses.

\subsubsection{Scope of PhysBench Transfer}
\label{supp:physbench_scope}

PhysBench is used purely as an out-of-distribution test; no PhysBench data enter \texttt{APT-Tune}.
Its four official sub-tasks aggregate over mechanical, chemical, optical, and thermal phenomena, which the current 14-type taxonomy covers unevenly by construction.
The reported gains are therefore evidence of transfer within macroscopic mechanics---consistent with Dynamics, the sub-task closest to APT semantics, showing the largest improvement---and not evidence of transfer to chemical, thermal, optical, or electromagnetic reasoning.
A phenomenon-level breakdown that would separate these regimes is left to future work.

\begin{lstlisting}[style=promptstyle, caption={Evaluation-time APT-schema prompt.}]
You are given 16 frames sampled uniformly from a physics video.

Task:
  Enumerate all visible Atomic Physical Transitions (APTs) in temporal order.
  Return a JSON array.
  Each item must contain:
    type: one of the 14 APT labels
    t_ms: approximate transition time in milliseconds

Allowed APT labels:
  contact_init: free then contact
  contact_break: contact then free
  elastic_rebound: approach then retreat after collision
  inelastic_capture: moving object contacts and remains captured/resting
  free_fall_onset: supported then falling
  projectile_apex: ascending then descending
  landing: falling then supported
  static_breakaway: stationary then sliding
  sliding_arrest: sliding then stationary
  surface_transition: object moves from one surface/material region to another
  rotation_onset: non-rotating then rotating
  toppling: stable then unstable/falling over
  rotation_arrest: rotating then stationary
  settling: motion damps to stable rest

Output rules:
  Return JSON only.
  Do not explain.
  Do not include labels that are not visible.
  Use milliseconds for t_ms.

Example output:
[
  {"type": "free_fall_onset", "t_ms": 420},
  {"type": "contact_init", "t_ms": 830},
  {"type": "elastic_rebound", "t_ms": 910}
]
\end{lstlisting}
\section{Limitations, Broader Impact, Failure Cases, and Reproducibility Details}
\label{app:responsible}

\subsection{Limitations}
\label{app:limitations}

\paragraph{What APT-Bench is.}
\aptbench{} is a transition-level benchmark and data construction framework.
The human-labeled component measures whether a model can enumerate APT chains in existing CLEVRER and Physion++ videos.
The Isaac Lab component uses the same APT labels as scene construction targets, allowing us to efficiently generate APT-style clips whose physical parameters can be customized.
This combination is important because existing videos provide realistic multi-step chains, while simulation lets us target rare or controlled transitions without waiting for them to appear naturally.

\paragraph{Limitation: current physics coverage.}
The current APT taxonomy mainly covers rigid-body and force-mediated transitions: contact, collision, gravity, friction, and rotation.
This scope is deliberate, because these transitions are visually grounded, frame-localized, and easy to instantiate in controlled simulation.
However, it does not cover all forms of physical change.
Fluid motion, deformable-body dynamics, thermal processes, material fracture, granular media, electromagnetism, and long-horizon causal interactions are outside the present data.
Future work will expand the taxonomy and generation pipeline to cover these broader APT families.

\paragraph{Limitation: causal-structure recovery is not causal reasoning.}
\aptbench{} scores the type and timestamp of each transition, where the type is a canonical name for a mechanism--state-change pair.
It therefore measures whether a model recovers a causally structured representation---which mechanism becomes active, in which direction the state changes, and when---and not whether the model can answer interventional or counterfactual queries, produce free-form mechanism explanations, or predict the effect of altering the scene.
These are distinct capabilities, and we make no claim about the latter.

\paragraph{Limitation: multi-object density.}
An APT is defined locally, over an entity, material region, or interacting pair, so a scene-level chain can already interleave transitions belonging to several entities and relations, and the present data do contain multi-object processes such as concurrent falling, landing, contact, and settling.
The benchmark nonetheless does not systematically stress \emph{dense} multi-object dynamics.
Most clips contain 2--3 APT-relevant moving or interacting objects, with at most 5, and only a small fraction of generated scenes contain as many as 7 total objects.
Simultaneous contacts among many bodies, severe mutual occlusion, and long interaction cascades are therefore under-represented, and we regard them as the most important extension of the current data rather than as settings the present results cover.

\paragraph{Limitation: annotation reliability reporting.}
Every trial was independently annotated by two of five trained annotators and adjudicated by a third before admission, but because adjudication was applied before records were committed we did not retain the pre-adjudication labels required to compute an inter-annotator agreement coefficient.
We therefore report the procedure rather than a $\kappa$ value.
Similarly, the GT-trace parser was calibrated iteratively against the human anchor and is used only for simulator-derived training data, never for the human-labeled test ground truth, so we do not report a held-out human-versus-parser agreement statistic either.
A dedicated double-annotation and parser-validation study is left to a future dataset release.

\paragraph{Limitation: choice of training objective.}
Our results establish that APT chains are effective structured supervision, not that supervised fine-tuning is the best way to consume them.
We did not evaluate timestamp-aware regression losses or reinforcement-learning objectives, both of which are plausible refinements; Section~\ref{sec:ablation} explains why they are non-trivial to formulate fairly here and why the observed error profile (38\,ms median timing error against 38.1\% recall) suggests limited headroom for timing-specific losses.
Objective optimality is outside the scope of this work.

\paragraph{Limitation: real-world APT localization.}
\aptbench{} is human-annotated on controlled physical videos, and our evidence for generalization beyond that distribution is indirect: PhysBench is a real-world OOD benchmark never used in training, and \texttt{APT-Tune} improves all six backbones on every PhysBench sub-task.
That is transfer evidence, not a direct evaluation of APT chain localization on unconstrained real-world video, which would require a new human-annotated in-the-wild split.

\paragraph{Limitation: annotation and simulation gap.}
Human labels provide semantic validity for real rendered videos, while Isaac Lab provides efficient controllability.
The two sources are complementary but not identical.
Some simulated clips may simplify visual texture, occlusion, or camera motion, while some human-labeled clips contain ambiguous boundaries that are hard to reproduce exactly in simulation.
Future releases should report the two sources separately and study how much generated APT-style data improves transfer to real videos.

\paragraph{Future directions.}
APT labels can serve as both evaluation units and generation specifications.
This suggests a scalable loop: define new APT labels, generate targeted clips in simulation, human-check boundary validity, and use the resulting examples to improve VLM training.
We view this label-first data construction process as the most direct path toward broader physics coverage.

\subsection{Broader Impact}
\label{app:broader_impact}

Our work has potentially positive impacts in settings that require physically grounded video understanding, including robotics, embodied agents, scientific video analysis, education, and safety-oriented evaluation of multimodal models.
By exposing transition-level physical failures that clip-level QA can miss, \aptbench{} can help diagnose brittle reasoning behavior and encourage models that are more faithful to visible causal structure rather than surface correlations.

Potential negative impacts are comparatively limited, because the benchmark focuses on short physical transitions in synthetic and controlled videos rather than personal, medical, or high-stakes human data.
Still, improved physical video understanding could indirectly strengthen systems used for automation or surveillance, so we emphasize transparent reporting, controlled benchmarking, and release of evaluation protocols together with limitations.
We do not release any high-risk generative model or scraped personal dataset as part of this work.

\subsection{Release, Statistical Reliability, and Compute Details}
\label{app:repro_compute}

\paragraph{Released assets and documentation.}
We release the new \apt{} dataset assets together with annotation guidelines, the multi-format SFT mixture, evaluation scripts, and LoRA training recipes.
We also credit the original sources used in our pipeline, including CLEVRER, Physion++, Something-Something v2, PhysBench, VideoPhy2, Qwen3-VL, and InternVL3.5, and follow the respective public licenses and terms of use of those assets.

\paragraph{Training and evaluation details.}
The main fine-tuning settings, prompts, frame sampling scheme, matching tolerance, and data splits are given in Section~\ref{sec:method}, Section~\ref{sec:experiments}, and Appendix~\ref{supp:evaluation_protocol}.
For the principal fine-tuning comparisons, we report averages over 3 independent runs with different random seeds; when variation is shown, it reflects across-run variability.
In addition, we report paired bootstrap significance tests on the 249 test trials with $n{=}1{,}000$ resamples.

\paragraph{Data collection compute.}
For simulator-based data construction, each \emph{generated} clip is simulated for approximately 20\,s of scene time at roughly real-time speed.
This 20\,s figure describes the Isaac Lab generation budget only and should not be read as the length of the evaluation clips, which average 9.77\,s (Table~\ref{tab:source_stats}).
Using 128 parallel Isaac Lab environments on a single RTX~4090, the idealized wall-clock simulator time is therefore about $20N/128$ seconds for $N$ videos.
For the released 1,246-trial split, this corresponds to approximately 195\,s of pure simulated time, excluding environment reset, rendering, trace export, and disk I/O overhead; in practice, total collection time is of the same order but somewhat larger.

\paragraph{Training compute.}
All LoRA fine-tuning runs use a single NVIDIA A100 GPU.
Representative end-to-end training times are approximately 12 hours for the 2B backbone, 15 hours for the 4B backbone, and 20 hours for the 8B backbone.
These numbers cover the reported LoRA training jobs and do not include unsuccessful exploratory runs.

\end{document}